\documentclass{article} % For LaTeX2e
\usepackage{iclr2026_conference,times}

\usepackage{amsmath,amsfonts,bm}

\def\eqref#1{equation~\ref{#1}}
\def\1{\bm{1}}

\DeclareMathAlphabet{\mathsfit}{\encodingdefault}{\sfdefault}{m}{sl}
\SetMathAlphabet{\mathsfit}{bold}{\encodingdefault}{\sfdefault}{bx}{n}

\usepackage{xcolor}
\usepackage{colortbl}
\usepackage{booktabs}
\usepackage{graphicx}
\usepackage{array}
\usepackage{hyperref}
\usepackage{url}
\usepackage{booktabs}
\usepackage{graphicx}
\usepackage{array}
\usepackage{capt-of}% or 
\usepackage{caption}
\usepackage{multirow}
\usepackage{pifont}
\usepackage{tikz}
\usepackage{wrapfig}
\usepackage{longtable}
\usepackage{booktabs}
\usepackage{multirow}
\usepackage{fontawesome5}
\usepackage{subcaption} % REQUIRED for \begin{subtable}
\usepackage{graphicx}   % For \resizebox
\usepackage{placeins}
\usepackage{marvosym}
\usepackage[dvipsnames]{xcolor} 
\usepackage{xcolor}
\usepackage{colortbl}
\usepackage{etoc}

\definecolor{tablered}{RGB}{230, 57, 70} % Red for Best
\definecolor{tableblue}{RGB}{29, 53, 87} % Blue for Best Open Source

\makeatletter
\DeclareRobustCommand\onedot{\futurelet\@let@token\@onedot}
\def\@onedot{\ifx\@let@token.\else.\null\fi\xspace}

\makeatother

\definecolor{deepgreen}{RGB}{0,100,0} % 深绿色 (DarkGreen)
\newcommand{\cmark}{\textcolor{deepgreen}{\checkmark}}

\definecolor{deepred}{RGB}{189,20,0}
\newcommand{\xmark}{\textcolor{deepred}{\ding{55}}}

\definecolor{amber}{RGB}{255,191,0} % 亮橙黄
\definecolor{deeporange}{RGB}{255,140,0} % 偏深橙黄

\newcommand{\omark}{\textcolor{deepgreen}{\checkmark}}

\usepackage{tcolorbox}

\newtcolorbox{takeawaybox}{
  colback=blue!5,        % 淡蓝背景
  colframe=blue!40,      % 浅蓝边框
  coltitle=black,
  boxrule=0.6pt,
  arc=3pt,
  left=8pt, right=8pt, top=6pt, bottom=6pt
}

\definecolor{red}{rgb}{1,0,0}
\newcommand{\revise}{}

\title{XModBench: Benchmarking Cross-Modal Capabilities and Consistency in Omni-Language Models}

\author{
Xingrui Wang\textsuperscript{1,2}, 
Jiang Liu\textsuperscript{1\Letter}, 
Chao Huang\textsuperscript{1,3}, 
Xiaodong Yu\textsuperscript{1}, 
Ze Wang\textsuperscript{1} \\
\textbf{Ximeng Sun}\textsuperscript{1},
\textbf{Jialian Wu}\textsuperscript{1},
\textbf{Alan Yuille}\textsuperscript{2},
\textbf{Emad Barsoum}\textsuperscript{1},
\textbf{Zicheng Liu}\textsuperscript{1} \\
\textsuperscript{1}Advanced Micro Devices \quad
\textsuperscript{2}Johns Hopkins University \quad
\textsuperscript{3}University of Rochester
}

\definecolor{lightred}{RGB}{255, 207, 206}      % 0-30: Light red
\definecolor{lightorange}{RGB}{255, 240, 201}   % 30-60: Light orange  
\definecolor{lightgreen}{RGB}{234, 255, 209}    % 60-90: Light green
\definecolor{lightblue}{RGB}{209, 244, 255}     

\newcommand{\perfcell}[1]{%
  \ifdim#1pt<30pt
    \cellcolor{lightred}#1%
  \else\ifdim#1pt<60pt
    \cellcolor{lightorange}#1%
  \else\ifdim#1pt<90pt
    \cellcolor{lightgreen}#1%
  \else
    \cellcolor{lightblue}#1%
  \fi\fi\fi
}

\hypersetup{
    colorlinks = true,
    urlcolor   = magenta
}

\iclrfinalcopy % Uncomment for camera-ready version, but NOT for submission.
\begin{document}

\maketitle

\vspace{-3em}
% \begin{center}
% \href{https://xingruiwang.github.io/projects/XModBench/}{\textcolor{magenta}{https://xingruiwang.github.io/projects/XModBench/}}
% \end{center}

\begin{center}
\vspace{0.4em}
\faGlobe\ \href{https://xingruiwang.github.io/projects/XModBench/}{\textcolor{magenta}{Project Page}}
\hspace{0.8em}
\href{https://huggingface.co/datasets/RyanWW/XModBench}{
  \raisebox{-0.2ex}{\includegraphics[height=0.9em]{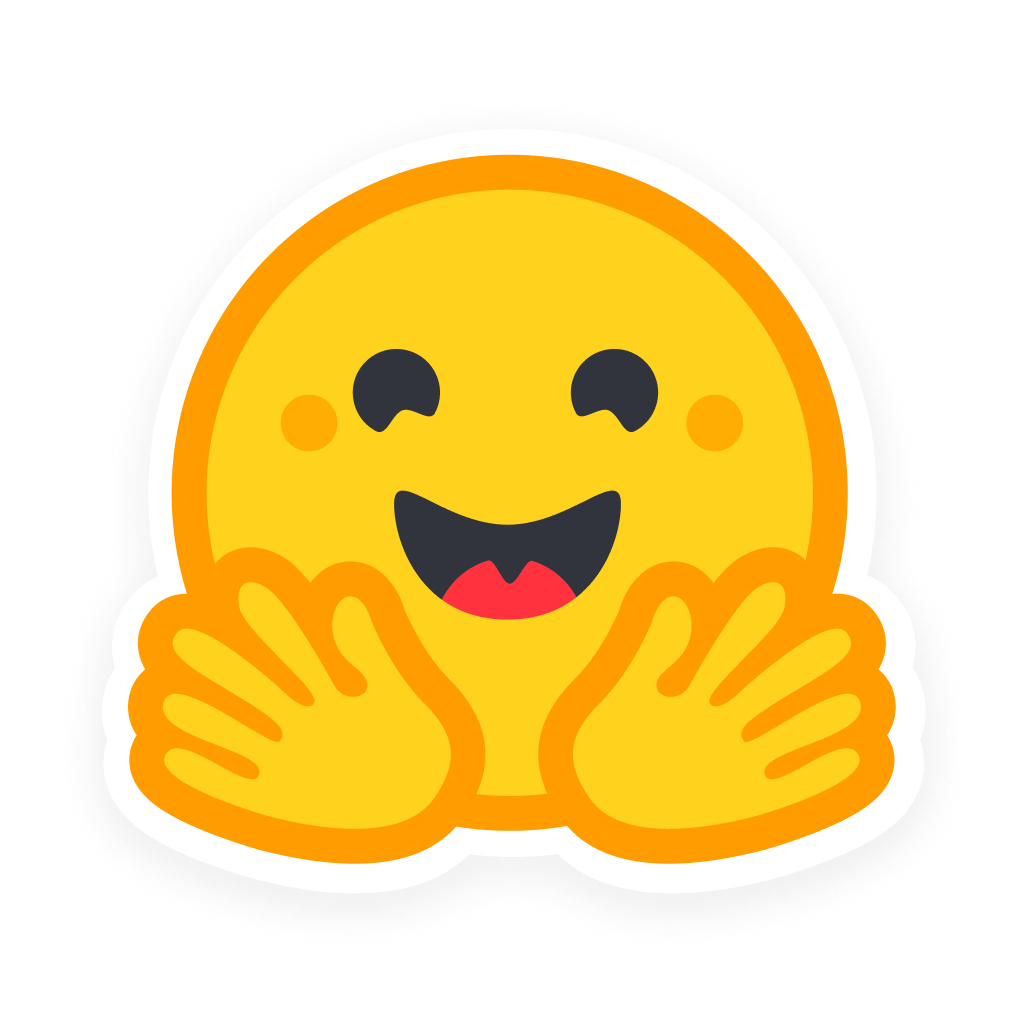}}\,
  Dataset Card
}
\hspace{0.8em}
\href{https://github.com/XingruiWang/XModBench}{
  \raisebox{-0.2ex}{\includegraphics[height=0.9em]{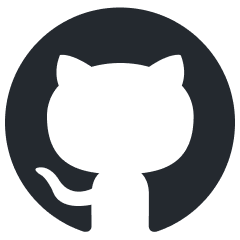}}\,
  Code
}
\end{center}

\begingroup
\renewcommand\thefootnote{}%
\footnotetext{\textsuperscript{\Letter} Corresponding author}
\addtocounter{footnote}{-1}%
\endgroup

\begin{figure}[h]
  \centering
    \includegraphics[width=0.92\linewidth]{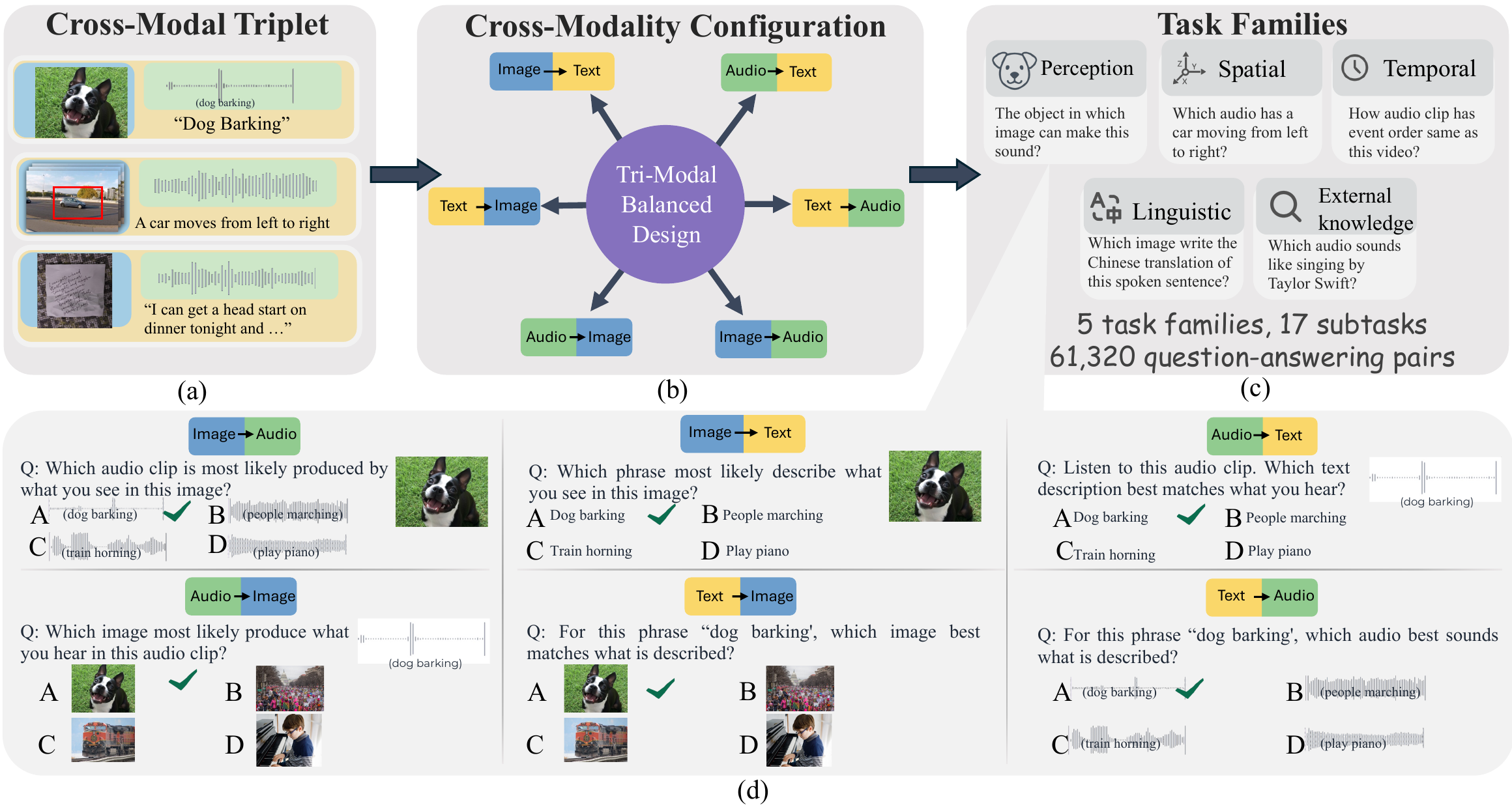}
    % \vspace{-20pt}
\caption{Overview of \textbf{XModBench}. (a) Instances are built from aligned text–image–audio triplets; (b) instantiated into six modality configurations by permuting context and candidate modalities; (c) spanning five domains with 17 subtasks and 61,320 question–answer pairs; and (d) illustrated with example multiple-choice questions under balanced modality settings.}
  \label{fig:overview}
\end{figure}

\begin{abstract}
Omni-modal large language models (OLLMs) aim to unify audio, vision, and text understanding within a single framework. 
While existing benchmarks primarily evaluate general cross-modal question-answering ability, it remains unclear whether OLLMs achieve modality-invariant reasoning or exhibit modality-specific biases. 
We introduce \textbf{XModBench}, a large-scale tri-modal benchmark explicitly designed to measure cross-modal consistency. 
XModBench comprises \textbf{61,320} multiple-choice questions spanning \textbf{five task families} and systematically covers all \textbf{six modality compositions} in question–answer pairs, enabling fine-grained diagnosis of an OLLM’s modality-invariant reasoning, modality disparity, and directional imbalance. 
Experiments show that even the strongest model, Gemini~2.5~Pro, 
(i) struggles with spatial and temporal reasoning, achieving less than 60\% accuracy, 
(ii) reveals persistent modality disparities, with performance dropping substantially when the same semantic content is conveyed through audio rather than text, and 
(iii) shows systematic directional imbalance, exhibiting lower consistency when vision serves as context compared to text. 
These findings indicate that current OLLMs remain far from truly modality-invariant reasoning, and position \textbf{XModBench} as a fundamental diagnostic tool for evaluating and improving cross-modal competence.
All data and evaluation tools will be available at \url{https://github.com/XingruiWang/XModBench}.

\end{abstract}

\section{Introduction}

Omni-modal large language models (OLLMs) integrate text, vision, and audio into a unified reasoning framework~\citep{comanici2025gemini,xu2025qwen2,xing2025echoink,su2023pandagpt,fu2025vita,cheng2024videollama,zhong2025omni}. 
However, despite impressive advancements and expanded modality coverage, a key question remains: do these models reason in a truly modality-invariant manner, or do they still exhibit systematic biases tied to specific input modalities?
For humans, cross-modal integration is typically seamless, yet it remains unclear whether OLLMs demonstrate comparable consistency. 
When the same semantic content is presented in different forms—spoken audio, written text, or visual images—do models still converge on the same correct answer? 
We refer to this property as \textit{cross-modal consistency}: the ability to maintain stable predictions regardless of input modality, thereby demonstrating reasoning over shared semantic representations rather than relying on modality-specific cues.
Although directly diagnosing whether current OLLMs achieve this goal is non-trivial, 
we can evaluate them through carefully designed benchmarks that expose inconsistencies. 
For instance, by posing semantically identical questions under different modality settings, we can test whether predictions diverge across modalities — an indicator of reliance on surface-level patterns rather than genuine modality-invariant reasoning.

Recent benchmarks have taken promising steps toward evaluating OLLMs, particularly through audio-visual tasks that reveal baseline cross-modality performance. 
Datasets such as Music AVQA~\citep{li2022learning}, AV-Reasoner~\citep{lu2025av}, and Pano-AVQA~\citep{yun2021pano} primarily probe fine-grained audio–visual reasoning, while broader efforts like AVQA~\citep{yang2022avqa}, WorldSense~\citep{hong2025worldsense}, AV-Odyssey Bench~\citep{gong2024av}, and OmniBench~\citep{li2024omnibench} expand to general multimodal understanding across diverse contexts. 
However, these benchmarks largely overlook whether models remain consistent across modalities. 
While other works~\citep{park2025assessing,zhang2024cross} attempt to assess modality consistency, they are restricted to the vision–text setting within vision–language models.

To address this gap, we introduce \textbf{XModBench}, a benchmark specifically designed to evaluate cross-modal consistency in omni-modal large language models.
We formulate all questions in a multiple-choice format, where each question naturally contains two components: 
(i) a \textit{context} describing an object or event, and 
(ii) a set of \textit{candidates} from which the model must select the correct one.
Unlike prior benchmarks that typically fix either the context or the choices to a single modality~\citep{yang2022avqa,li2024omnibench}, XModBench systematically covers all six cross-modal directions among audio, vision, and text (see Tab.~\ref{tab:benchmark_comparison}).
To ensure broad coverage and rigorous evaluation, \textsc{XModBench} spans five domains—perception, spatial reasoning, temporal reasoning, linguistic understanding, and external knowledge. 
We curate data across these domains through re-annotation, synthetic construction, and targeted web collection, ensuring both diversity and balance across modalities. 
The resulting benchmark comprises \textbf{61,320} multiple-choice question–answer pairs (10,220 unique instances), each instantiated in six modality configurations that preserve identical semantics across audio, visual, and textual forms. 
This enables both large-scale evaluation and fine-grained diagnosis of cross-modal consistency. 
An overview of the benchmark design is illustrated in Fig.~\ref{fig:overview}.

\begin{table}[!b]
\centering
\caption{Comparison of multimodal question-answering (QA) benchmarks by modality coverage, task domains, and modality consistency.}
\vspace{-5pt}
\label{tab:benchmark_comparison}
\resizebox{\textwidth}{!}{
\begin{tabular}{l|c|ccc|ccc|ccccc|c}
\toprule
\multirow{2}{*}{\textbf{Benchmark}} & \multirow{2}{*}{\textbf{\#Q}} & \multicolumn{3}{c}{\textbf{Context Modality}} & \multicolumn{3}{c|}{\textbf{Candidate Modality}} & \multicolumn{5}{c|}{\textbf{Task Domain}} & \multirow{2}{*}{\textbf{Mod. Consist.}} \\
\cmidrule(lr){3-5} \cmidrule(lr){6-8} \cmidrule(lr){9-13}
 &  & Text & Vision & Audio  & Text & Vision & Audio  & Percep. & Spatial & Temporal & Ling. & Ext. Know. &  \\
\midrule
MME Bench~\citep{fu2024mmecomprehensiveevaluationbenchmark} & 2,194  & \xmark &  \cmark & \xmark  & \cmark  &  \xmark &  \xmark &  \cmark &  \xmark & \xmark  &  \cmark & \cmark & \xmark \\
MMBench~\citep{liu2024mmbench} & 3,217  &  \xmark &  \cmark & \xmark  & \cmark   & \xmark &  \xmark &  \cmark  & \cmark   &  \xmark & \cmark   &  \cmark  & \xmark \\
OcrBench v2~\citep{fu2024ocrbench} & 10,000 &  \xmark &  \cmark & \xmark  & \cmark  & \xmark  & \xmark  &  \cmark &  \xmark &  \cmark & \xmark  &  \xmark & \xmark \\
SEED-Bench-2~\citep{li2024seed} & 24,371 & \cmark &  \cmark  &  \xmark & \cmark   & \cmark  & \xmark  &  \cmark  & \cmark   &  \cmark &  \cmark &  \cmark & \xmark \\
\midrule
AudioBench~\cite{wang2024audiobench} & 24,371 &  \xmark &  \xmark &  \cmark & \cmark  & \xmark  & \xmark  &  \cmark &  \xmark & \xmark  & \cmark  & \xmark  & \xmark \\
Audiopedia~\citep{li2022learning} & 45,867 & \xmark  & \xmark  &  \cmark &  \cmark & \xmark & \xmark  &  \xmark & \xmark  &  \xmark & \cmark  & \cmark  & \xmark \\
MMAU~\citep{sakshi2024mmau} & 10,000 &  \xmark &  \xmark& \cmark  &  \cmark & \xmark &  \xmark &  \cmark  &  \xmark & \xmark  & \cmark   &  \xmark & \xmark \\ 
\midrule
AVQA~\citep{yang2022avqa} & 57,335 &  \xmark & \cmark  & \cmark  & \cmark  &  \xmark & \xmark  &  \cmark &  \cmark &  \cmark &  \xmark & \xmark  & \xmark \\
Pano-AVQA~\citep{yun2021pano} & 51,700 &  \xmark & \cmark  &  \cmark & \cmark  &  \xmark &  \xmark &  \cmark & \cmark & \xmark  &  \xmark &  \xmark & \xmark \\
Music-AVQA~\citep{li2022learning} & 45,867 &  \xmark & \cmark  &  \cmark & \cmark  & \xmark  &  \xmark &  \cmark &  \cmark &  \cmark &  \xmark & \xmark  & \xmark \\
SAVE Bench~\citep{sun2024video} & 4,350  & \xmark  & \cmark  & \cmark  & \cmark  &  \xmark &  \xmark & \cmark  &  \xmark & \xmark  & \cmark  & \xmark  & \xmark \\
Video-MME~\citep{fu2025video} & 2,700  &  \xmark &  \cmark & \cmark  & \cmark  &  \xmark &  \xmark & \cmark  &  \cmark  & \cmark   &  \cmark  &  \cmark & \xmark \\
WorldSense~\citep{hong2025worldsense} &   3,172    & \xmark  &  \cmark  & \cmark   & \cmark   & \xmark   &  \xmark  &  \omark  &  \xmark  &  \xmark   & \xmark   &  \omark & \xmark \\
AV-Reasoner~\citep{lu2025av} &    1,027   & \xmark  &  \cmark  & \cmark   & \cmark   & \xmark   &  \xmark  &  \xmark  &  \xmark  &  \cmark  & \xmark   &  \xmark  & \xmark \\
AV-Odyssey Bench~\citep{gong2024av}  & 1,142  &  \xmark & \cmark  &  \cmark & \cmark  & \cmark  & \cmark  & \omark  &  \omark & \omark  &  \xmark & \omark  & \xmark \\
OmniBench~\citep{li2024omnibench} & 4,555  & \xmark  & \cmark  &  \cmark &  \cmark & \xmark  & \xmark  & \omark  & \xmark  & \xmark  &  \omark &  \xmark & \xmark \\ 
\midrule
\textbf{XModBench (Ours)} & 61,320 & \cmark  & \cmark  & \cmark  &  \cmark & \cmark  & \cmark  & \cmark  & \cmark  & \cmark  & \cmark  & \cmark  & \cmark \\
\bottomrule
\end{tabular}}
\end{table}

We systematically evaluate models on \textsc{XModBench}, going beyond overall accuracy to provide fine-grained diagnosis of cross-modal reasoning. 
Specifically, we analyze three complementary dimensions: 
(1) \textbf{Task competence}—by averaging over all six modality directions, we assess model performance across perception, spatial, temporal, linguistic, and knowledge tasks, yielding task-centric comparisons of multimodal competence; 
(2) \textbf{Modality disparity}—we measure consistency when the same question is posed in different modalities, where high variability signals reliance on modality-specific cues rather than shared semantic representations; and 
(3) \textbf{Directional imbalance}—we compare accuracy when context and candidate modalities are swapped, revealing asymmetries in cross-modal grounding. 

Our experiments show that current OLLMs fall short along all three axes. 
They perform strongly on perception and linguistic tasks (best models reach around 70\%), but degrade by 15–25 points on spatial and temporal reasoning. 
Performance also drops sharply whenever audio is involved, underscoring that auditory representations remain the weakest link. 
Finally, accuracy is consistently higher when text serves as the candidate modality, highlighting incomplete bidirectional alignment across modalities. 
Together, these findings demonstrate that today’s OLLMs remain far from achieving modality-invariant reasoning, underscoring the diagnostic value of \textsc{XModBench}.

In summary, \textsc{XModBench} makes the following key contributions:
\begin{enumerate}
    \item \textbf{Cross-modal consistency benchmark.} We present \textsc{XModBench}, the first tri-modal multiple-choice question-answering benchmark explicitly designed to evaluate cross-modal consistency, covering all six modality mappings among audio, vision, and text. 
    \item \textbf{Comprehensive coverage.} The benchmark spans five task families with 17 subtasks and 61,320 question–answer pairs, ensuring broad domain coverage and fine-grained diagnostics, while its balanced design enables fair assessment of modality-invariant reasoning. \vspace{-5pt}
    \item \textbf{Diagnostic metrics.} We introduce \emph{modality disparity} and \emph{directional imbalance} to directly measure robustness and bidirectional alignment across modalities. Our experiments reveal systematic weaknesses in current OLLMs, providing actionable insights for developing more modality-invariant architectures and training strategies.
\end{enumerate}

\vspace{-1em}
\section{Related Work}
\vspace{-8pt}

\looseness -1 \textbf{Multimodal Question Answering (QA) Benchmarks.} 
% A number of benchmarks have been developed to evaluate multimodal language models (MLMs). Grouped by modality composition, many works focus on the vision-text setting (covering both images and videos), such as MMEBench~\citep{yin2024survey}, MMBench~\citep{liu2024mmbench}, and SEED-Bench-2~\citep{li2024seed}. For audio–text evaluation, representative efforts include AudioBench~\citep{wang2024audiobench} and MMAU~\citep{sakshi2024mmau}.
% When combining audio and vision with text, a variety of benchmarks have emerged, including AVQA~\citep{yang2022avqa}, Music-AVQA~\citep{li2022learning}, Audiopedia~\citep{li2022learning}, Pano-AVQA~\citep{yun2021pano}, SAVE Bench~\citep{sun2024video}, WorldSense~\citep{hong2025worldsense}, AV-Reasoner~\citep{lu2025av}, AV-Odyssey~\citep{gong2024av}, OmniBench~\citep{li2024omnibench}, and DailyOmni~\citep{zhou2025daily}. Other recent works, such as ACVUBench~\citep{yang2025acvubench}, also extend evaluation to diverse multimodal combinations.
A number of benchmarks have been developed to evaluate multimodal large language models (MLLMs). 
Grouped by modality composition, \citet{yin2024survey}, \citet{liu2024mmbench}, and \citet{li2024seed} focus on the vision–text setting (covering both images and videos). 
For audio–text evaluation, representative efforts include \citet{wang2024audiobench} and \citet{sakshi2024mmau}. 
When combining audio and vision with text, a variety of benchmarks have emerged, including \citet{yang2022avqa}, \citet{li2022learning}, \citet{yun2021pano}, \citet{sun2024video}, \citet{hong2025worldsense}, \citet{lu2025av}, \citet{gong2024av}, \citet{li2024omnibench}, and \citet{zhou2025daily}. 
Other recent works, such as \citet{yang2025acvubench}, further extend evaluation to diverse multimodal combinations.
Despite their breadth, these benchmarks primarily emphasize coverage across tasks and modalities, while less attention has been paid to evaluating \emph{cross-modal consistency}—whether models produce stable answers when the same semantic content is expressed in different modality forms. Our work fills this gap by explicitly designing a benchmark centered on modality-invariant reasoning.

% \textbf{Interleaved benchmark}
% AURA: A Fine-Grained Benchmark and Decomposed Metric for Audio-Visual
% Reasoning
% Daily-Omni: Towards Audio-Visual Reasoning with Temporal Alignment across Modalities
% AV-Reasoner: Improving and Benchmarking Clue-Grounded Audio-Visual Counting for MLLMs

% \textbf{Modality Consistency}
\textbf{Cross-Modality Consistency.} 
Recent work has begun to investigate whether multimodal models behave consistently across modalities. \citet{park2025assessing} introduced the Modality Importance Score to quantify modality bias, which measures how much each modality contributes to answering questions in VideoQA. \citet{zhang2024cross} further proposed the notion of cross-modal consistency between text and image, defining a consistent model as one that applies the same internal reasoning to semantically identical inputs across modalities, thereby yielding consistent outcomes. In contrast, other studies, such as \citet{sung2024avhbench} and \citet{choong2024vidhal}, report instances of inconsistent audio-video reasoning, where models hallucinate non-existent sounds or visual signals, thereby exposing modality bias and cross-modal inconsistency.
While these efforts provide pioneering insights into cross-modal consistency, they are typically confined to specific modality pairs. Our work not only expands the scope to cover a broader range of modality combinations for state-of-the-art OLLMs, but also conducts a deeper analysis of their cross-modality reasoning behavior on a comprehensive task suite.
% \cite{guan2024hallusionbench}

% \input{main/03_benchmark2}
\section{XModBench: Comprehensive cross-modal balanced benchmark}
\vspace{-1em}

\looseness -1 We introduce \textbf{XModBench}, a comprehensive multiple-choice question-answering (QA) benchmark designed to evaluate the cross-modal capabilities and consistency of OLLMs across audio, vision, and text. 
A key feature of \textbf{XModBench} is its modality-balanced design, which creates six cross-modal variants of semantically identical questions to enable a controlled and fair evaluation of cross-modal capabilities and consistency (Sec.~\ref{sec:paired}). 
The benchmark offers extensive domain coverage through five task families and seventeen subtasks (Sec.~\ref{sec:tasks}), all built upon meticulously curated, high-quality, and diverse tri-modal data (Sec.~\ref{sec:data}).

\vspace{-0.7em}
\subsection{Benchmark Design}
\vspace{-0.7em}
\label{sec:paired}

The central objective of \textsc{XModBench} is to evaluate whether models preserve \emph{cross-modal consistency} when the same semantic content appears in different modalities. 
Each item is a four-choice multiple-choice question consisting of a \texttt{<context>} (question stem) and four \texttt{<candidates>} (answer options). 
By systematically permuting text (T), vision (V), and audio (A) across the \texttt{<context>} and \texttt{<candidates>}, we generate six modality configurations of the same question (see Fig.~\ref{fig:overview} (b) and (d)). 
This balanced design ensures that no single modality is privileged and enables consistent evaluation across all directions, which supports three diagnostic properties aligned with the goals of our benchmark:

\textbf{(1) Task competence.}
Since each task is instantiated uniformly across all modality pairs, we measure competence by averaging accuracy across all context–candidate configurations. 
This yields a fair estimate of a model’s overall capability for each task, independent of modality-specific biases.

\textbf{(2) Modality disparity.}  
By presenting semantically identical questions under different modality configurations, we keep the content fixed while varying only the modality. 
For example, to compare audio and vision, we examine cases where text provides the context with audio candidates (T$\mapsto$A) versus text with visual candidates (T$\mapsto$V), and similarly compare A$\mapsto$T against V$\mapsto$T settings. 
Differences in accuracy under these controlled comparisons reveal modality disparities, indicating the relative competence across different modalities.

\textbf{(3) Directional imbalance.}  
We examine inverse settings by swapping the modalities of context and candidates. 
For example, a model may perform well when vision serves as the context and text provides the options (V$\mapsto$T), but perform worse when the same semantic content is presented as a text context with visual candidates (T$\mapsto$V). 
Such differences indicate asymmetric grounding between the two modalities, and comparable asymmetries are also observed in the audio–text and audio–vision pair.

\begin{wrapfigure}{r}{0.45\textwidth}
  \centering
  \vspace{-35pt}
  \includegraphics[width=0.43\textwidth]{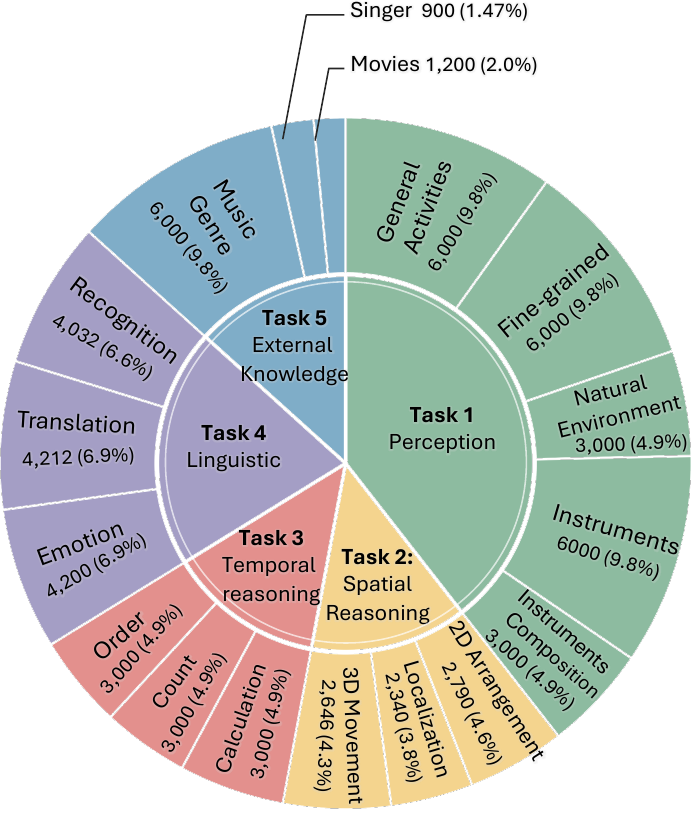}
  \caption{Distribution of XModBench's questions across five task families with specific subtasks.}
  \label{fig:distribution}
  \vspace{-5pt}
\end{wrapfigure}

\vspace{-0.8em}
\subsection{Task Taxonomy}
\vspace{-0.8em}
\label{sec:tasks}

XModBench covers five task families with seventeen subtasks, spanning perception, spatial reasoning, temporal reasoning, linguistic understanding, and external knowledge (see Fig.~\ref{fig:distribution}).  
Each task is formulated in the multiple-choice format and follows the modality-balanced configuration described in Section~\ref{sec:paired}: a \texttt{<context>} is drawn from one modality and four \texttt{<candidates>} from another.  
In this section, we detail the design of these subtasks and specify how each instance is instantiated across modalities within every task.

\begin{figure}[t]
  \centering
    \includegraphics[width=0.95\linewidth]{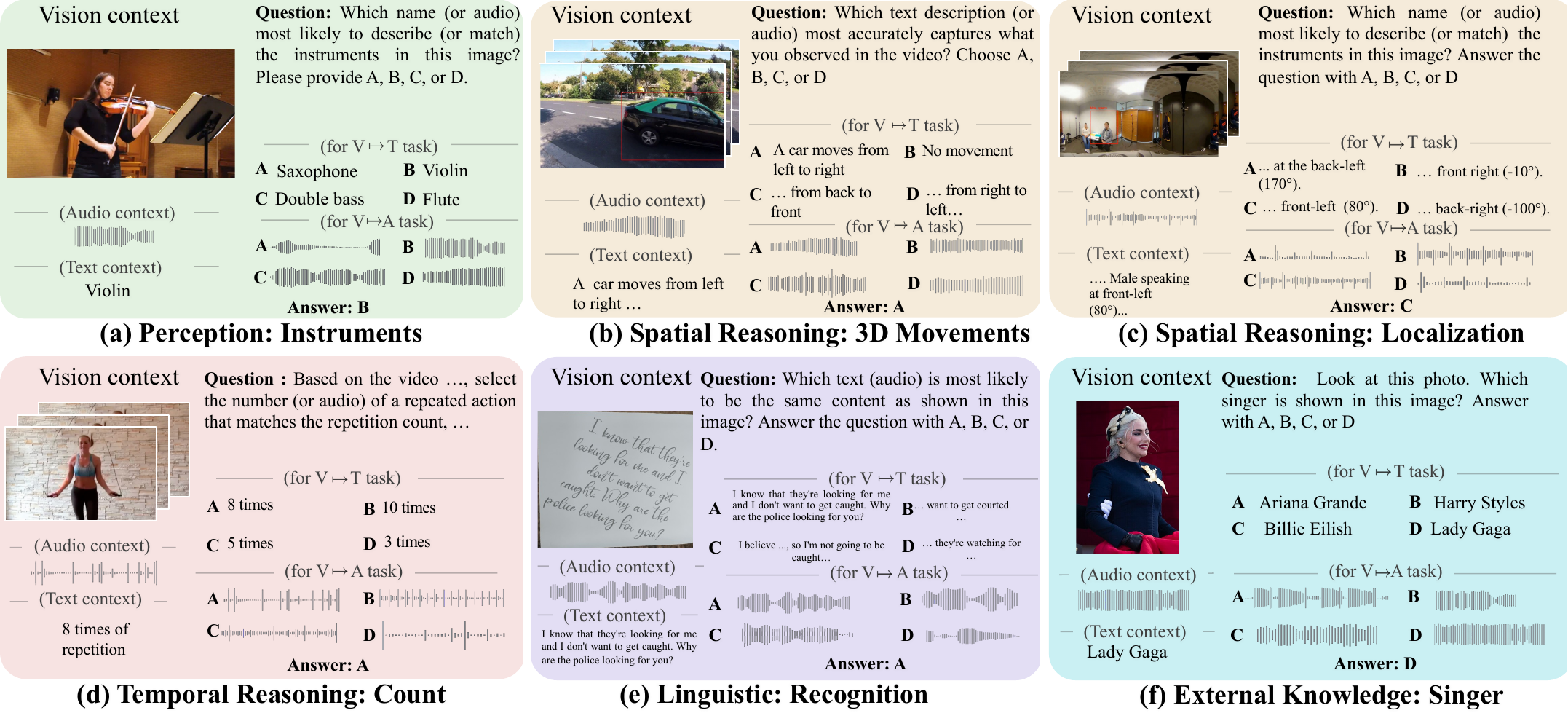}
\caption{XModBench task examples. We show sample questions from five subtasks in the benchmark. Each question includes possible contexts from different modalities, and for the vision-context example, the candidates are given in either text or audio.}\vspace{-1em}
  \label{fig:example}
\end{figure}

\textbf{Task 1. Perception.}  
This task evaluates whether models can recognize the same object, activity, or scene across modalities. 
For example, a barking dog may appear as an image, as its sound, or as the text description “dog barking.”  
Here, {visual} inputs are images, {audio} inputs are recordings of corresponding sounds, and {text} inputs are short labels or phrases. 
The data are drawn from diverse domains, including human activities, animal behaviors, musical instruments, and natural environments.

We divide perception into several subtasks. 
\textbf{General activity} recognition mixes candidates from diverse domains to test broad semantic alignment, while \textbf{fine-grained activity} recognition restricts candidates to a single domain (e.g., animal species or instrument types), thereby increasing difficulty and requiring precise discrimination. 
We further design domain-specific subtasks to capture unique challenges: recognizing \textbf{natural environments} (e.g., rainfall, wind, fire), distinguishing \textbf{instruments} (e.g., violin, bass, cello), and identifying \textbf{instrument compositions} where multiple instruments are played together (e.g., violin and bass, or cello and flute). 
Illustrative examples are shown in Fig.~\ref{fig:overview}(d) and Fig.~\ref{fig:example}(a).

\textbf{Task 2. Spatial reasoning.}  
This task evaluates whether models can interpret object positions and motion in 2D and 3D space, which is an important factor in vision–language models~\citep{chen2024spatialvlm}. 
We extend this ability to the omni-modal setting and design three subtasks. 
The first is \textbf{2D arrangement}, where the model determines the left–right order of objects such as musical instruments. 
Visual inputs are images of ordered layouts, audio inputs are stereo recordings with left–right cues, and text inputs describe the relative arrangement; distractors are generated by swapping or permuting positions. 
The second subtask, \textbf{3D localization}, using panoramic videos from~\citet{shimada2023starss23}, requires identifying the orientation of events in video frames, spatialized audio, and short textual descriptions (e.g., “a man speaking from the front-left”); distractors are produced by shifting the same scene to nearby but incorrect directions through camera or audio rotation. 
The third subtask, \textbf{3D movement understanding}, focuses on motion directions such as left–right or front–back, instantiated with street-view or action videos, spatialized audio of approaching or receding sounds, and textual trajectory descriptions~\citep{fuentes2022urban}; distractors are clips with incorrect motion patterns or mismatched vehicle types. 
Examples for the 3D movement and localization tasks are shown in Fig.~\ref{fig:example}(b) and (c), respectively.

\textbf{Task 3. Temporal reasoning.}  
This task evaluates whether models can understand \textbf{event order} and \textbf{frequency} across time in video and audio. 
We design three subtasks. 
The first is \textbf{temporal order}, where models infer the correct sequence of events from muted video segments, audio clips, or textual descriptions and align them across modalities. 
The second, \textbf{temporal counting}, requires recognizing the number of repeated actions such as tennis hits, jumps, or drum beats, with distractors differing in count. 
For example, a video may show a tennis player hitting the ball three times, and the model must select the audio clip with exactly three hits or the text “3 times.” 
The third, \textbf{temporal calculation}, extends counting by applying simple arithmetic to the repetition number. 
For instance, if a video shows a person jumping three times and the query applies $2 \times \text{count}$, the correct answer should correspond to six repetitions, given either as an audio clip with six jumps or as the text “6 times.” 
An example of the temporal counting task is in Fig.~\ref{fig:example}(d).

\textbf{Task 4. Linguistic understanding.}  
This task covers recognition of linguistic content and interpretation of affective meaning. 
While prior work separates OCR for vision and ASR for audio~\citep{fu2024ocrbench,wang2024audiobench}, XModBench unifies them in a cross-modal setting. 
We design three subtasks. 
The first, \textbf{linguistic recognition}, focuses on transcribing text from images, audio, or phrases; correct candidates require word-level precision, while distractors differ by only one or two words (see Fig.~\ref{fig:example}(e)). 
The second, \textbf{translation}, evaluates English–Chinese translation across modalities, with distractors introducing subtle shifts such as antonyms, degree modifiers (e.g., “very” → “a little”), or small changes in numbers and entities. 
The third, \textbf{emotion classification}, targets affective understanding in dialogue: audio inputs are spoken conversations, visual inputs are muted video clips with transcripts, and candidates represent emotions such as joy, sadness, or anger, with distractors drawn from closely related categories.

\textbf{Task 5. External knowledge.}  
Beyond perceptual and reasoning skills, some tasks require linking multimodal content with world knowledge. 
We design three subtasks. 
The first, \textbf{movie recognition}, presents audio clips from trailers, visual posters, or short text descriptions of the plot, with candidates drawn from films of similar genres or storylines. 
The second, \textbf{music genre classification}, uses album covers, short audio clips, or textual genre labels, with distractors from closely related genres (e.g., “jazz” vs. “blues”). 
The third, \textbf{singer identification}, provides names, portrait images, or audio clips of songs, with distractors sampled from artists of similar musical styles (see Fig.~\ref{fig:example}(f)).

\vspace{-0.8em}
\subsection{Data Curation}
\vspace{-0.8em}
\label{sec:data}

\looseness -1 The construction of XModBench follows a three-stage pipeline. 
We begin by collecting large-scale text–vision–audio triplets across all task domains, then generate task-specific multiple-choice questions, and finally apply both automated filtering and human verification to ensure quality and consistency.  

\textbf{Cross-modal data collection.}  
We curate a large corpus aligned across vision, audio, and text by combining three sources:  
(i) re-annotated and extended data from existing multimodal datasets, such as adapting VGG-Sound for perception tasks or STARSS23~\citep{lee2022sound,shimada2023starss23} for spatial reasoning;  
(ii) synthetic or model-generated content to cover missing modalities, for example generating speech audio with FireRedTTS~\citep{guo2025fireredtts} or producing rendered text images for translation tasks; and  
(iii) web-collected samples for domains not well represented in public resources, such as singer portraits and songs for the \emph{Singer Identification} task or trailers and posters for \emph{Movie Recognition} from public YouTube videos.  
This design ensures both coverage and balance across all five task families. 
Detailed sources and processing procedures are described in Appendix~\ref{app:data}.

\textbf{Question candidate generation.}
To ensure the correctness of both the generated questions and answers, we first construct task-specific multiple-choice templates using our annotated tri-modality data. The question descriptions are then refined by GPT-5~\citep{openai2025gpt5} solely to improve language fluency and stylistic diversity. Importantly, this refinement does not introduce any new information or alter the underlying semantics.
Each question is instantiated with a context and four candidates under the modality-balanced configuration, ensuring consistent evaluation across all modality directions. 
Distractors are created to be semantically challenging but unambiguous, while templates are diversified with both human-written prompts and LLM-assisted variations.  

\textbf{LLM filtering and human-in-the-loop verification.}  
To guarantee data quality at scale, we first adopt foundation models~\citep{openai2025gpt5,comanici2025gemini} to filter out low-quality or ambiguous samples. 
Human annotators then double-check the filtered results to ensure accuracy. 
After questions are constructed, an internal round of testing is conducted by annotators, who resolve ambiguities and validate correctness. 
Feedback from this process is used to regenerate and retest questions until high-quality items are obtained.

Overall, this pipeline yields a high-quality benchmark with diverse and well-aligned multimodal content. 
More detailed descriptions of dataset sources, generation strategies, and signal-processing techniques are provided in Appendix~\ref{app:data}.

\vspace{-0.5em}
\section{Experiments}
\vspace{-0.5em}
\subsection{Baselines}
\vspace{-0.5em}
% We evaluate XModBench on a diverse set of recent omni-language and multimodal models. The Gemini family \citep{comanici2025gemini,team2024gemini} represents close source frontier systems with strong vision–language–audio integration and we test 4 models from Gemini 2.5 pro to Gemini 1.5 pro. For the open-source model, Qwen2.5-Omni \citep{xu2025qwen2} is a large-scale multilingual model with multimodal extensions. The other latest open source model include Baichuan Omni 1.5 and Echoink-R1 \citep{li2025baichuan,xing2025echoink} are omni-models focusing on cross-lingual and cross-modal alignment. We also include omni-modal model VideoLLaMA 2 \citep{cheng2024videollama}, Vita \citep{fu2025vita}, Unified-IO 2 \citep{lu2024unified} series (large,XL and XXL variants) and PandaGPT \citep{su2023pandagpt}.
% \subsection{Baselines}

We evaluate \textsc{XModBench} on a diverse set of recent omni-modal large language models. 
The \textbf{Gemini series}~\citep{comanici2025gemini,team2024gemini} represents state-of-the-art closed-source omni-modal models, and we include multiple variants ranging from Gemini~1.5~Pro to Gemini~2.5~Pro. 
Note that OpenAI APIs do not currently support processing audio and visual modalities jointly within a single query; therefore, we omit the GPT series from our evaluation. 
For open-source models, we include the latest \textbf{Qwen2.5-Omni}~\citep{xu2025qwen2}, \textbf{Baichuan~Omni~1.5}~\citep{li2025baichuan}, and \textbf{EchoInk-R1}~\citep{xing2025echoink}. 
Additional open-source omni-modal baselines include \textbf{VideoLLaMA~2}~\citep{cheng2024videollama}, \textbf{VITA}~\citep{fu2025vita}, the \textbf{Unified-IO~2} series (Large, XL, and XXL variants)~\citep{lu2024unified}, and \textbf{PandaGPT}~\citep{su2023pandagpt}. 
Together, these models represent a broad spectrum of both closed- and open-source OLLMs.

% Together, these baselines span close-source and open-source systems, covering different architectures, training paradigms, and deployment scales.
% \subsection{Prompt}

\vspace{-0.5em}
\subsection{Model Performances}
\vspace{-0.5em}

\begin{table*}[!t]
\centering
\small
\caption{Results on \textbf{XModBench}. We report (a) the performance under different input modalities across the full benchmark, and (b) the summary of average accuracy for each of the 5 task families. The highest scores are \textbf{bolded}, and the second highest are \underline{underlined}.}

% Scores are color-coded as \colorbox{lightred}{$<30$}, \colorbox{lightorange}{30–60}, \colorbox{lightgreen}{60–90}, \colorbox{lightblue}{$\geq 90$}.}}
\label{tab:multimodal_performance_lite}
\resizebox{\textwidth}{!}{
    \begin{tabular}{l|ccccc|ccccccc|c}
    \toprule
    \multirow{2}{*}{\textbf{Model}} & \multicolumn{5}{c|}{\textbf{Accuracy on 5 Task Families}} & \multicolumn{7}{c|}{\textbf{Modality Configuration}} & \multicolumn{1}{c}{\textbf{{Avg.}}} \\
    & \textbf{Perc.} & \textbf{Spat.} & \textbf{Temp.} & \textbf{Ling.} & \textbf{Knwl.} & \textbf{A $\mapsto$ T} & \textbf{A $\mapsto$ V} & \textbf{T $\mapsto$ A} & \textbf{T $\mapsto$ V} & \textbf{V $\mapsto$ A} & \textbf{V $\mapsto$ T} & \textbf{\textit{Std.}} &  \\
    \midrule 
    No Context & 25.5 & 24.8 & 24.9& 24.7 & 25.5 & 25.1 & 24.3 & 25.4 & 24.8 & 25.3 & 25.7 & 0.4 & 25.1 \\
    \midrule
    Qwen2.5-VL & 91.3 & 51.4 & 40.9 & 84.1 & 77.2 & - & - & - & {60.1} & - & {74.7} & - & {67.4} \\
    Intern3.5-VL & 87.2 & 42.7 & 41.4 & 75.0 & 68.7 & - & - & - & {49.7} & - & {73.7} & - & {61.7} \\
    \midrule
    PandaGPT & {24.6} & {25.7} & {24.4} & {25.5} & {23.1} & {24.5} & {25.0} & {23.8} & {25.2} & {24.5} & {25.1} & {0.5} & 24.7 \\
    Unified-IO 2 & {36.1} & {23.6} & {23.8} & {30.4} & {26.8} & {28.9} & {24.0} & {25.4} & {32.0} & {25.7} & {32.7} & {3.7} & 28.1 \\
    Unified-IO 2 XL & {42.2} & {25.0} & {26.1} & {30.8} & {29.5} & {33.3} & {27.0} & {27.1} & {32.9} & {26.5} & {37.4} & {4.5} & 30.7 \\
    Unified-IO 2 XXL & {43.7} & {28.3} & {27.7} & {31.2} & {34.0} & {37.4} & {25.0} & {31.2} & {37.8} & {26.7} & {39.9} & {6.3} & 33.0 \\
    VideoLLaMA 2 & {45.7} & {33.9} & {29.2} & {36.7} & {36.8} & {48.6} & {26.0} & {25.7} & {26.5} & {25.2} & {66.8} & {17.4} & 36.5 \\
    VITA & {34.8} & {34.0} & {29.4} & {46.1} & {32.6} & {40.2} & {26.0} & {29.8} & {26.8} & {29.9} & {59.3} & {12.8} & 35.4 \\
    Baichuan Omni 1.5 & {58.9} & {34.9} & {30.0} & {62.8} & {56.7} & {47.8} & {{35.8}} & {40.5} & {56.2} & {38.6} & {73.0} & {14.0} & 48.7 \\
    EchoInk-R1 & {75.8} & {36.6} & {37.1} & {73.3} & {73.3} & \underline{{64.6}} & {45.9} & \underline{{56.4}} & {60.9} & {49.9} & {77.6} & {11.3} & {59.2 }\\
    Qwen2.5-Omni & {75.5} & {38.4} & {32.3} & {74.1} & {72.8} & {62.0} & {48.0} & {55.4} & {59.6} & {{50.5}} & {76.3} & {10.1} & 58.6 \\ 
    \midrule
    Gemini 1.5 Pro & {56.2} & {40.1} & {37.1} & {72.6} & {69.4} & {52.4} & {38.2} & {48.6} & {70.4} & {40.7} & {79.9} & {16.7} & 55.0 \\
    Gemini 2.0 Flash & {66.2} & {48.4} & {44.8} & {70.2} & {78.1} & \underline{{63.7}} & \underline{{49.0}} & {52.2} & {71.5} & {47.6} & {85.2} & {15.2} & 61.2 \\
    Gemini 2.5 Flash & {66.1} & {48.0} & {48.6} & {73.1} & {82.8} & {62.6} & {51.2} & {55.1} & \underline{{75.7}} & \underline{{51.9}} & \underline{{86.0}} & {14.2} & \underline{63.7} \\
    Gemini 2.5 Pro & \textbf{{75.9}} & \textbf{{50.1}} & \textbf{{60.8}} & \textbf{{76.8}} & \textbf{{89.3}} & \textbf{{71.0}} & \textbf{{58.9}} & \textbf{{64.4}} & \textbf{{79.8}} & \textbf{{60.8}} & \textbf{{88.6}} & {11.7} & \textbf{70.6} \\
    \midrule
    Human & {91.0} & {89.7} & {88.9} & {93.9} & {93.9} & {92.4} & 91.5 & 91.1 & 91.8 & 86.4 & 95.6 & 3.0 & 91.5 \\
    \bottomrule
    \end{tabular}
}
\end{table*}

Table~\ref{tab:multimodal_performance} reports results across five task families and six cross-modal directions (Audio $\mapsto$ Text, Audio $\mapsto$ Vision, Text $\mapsto$ Audio, Text $\mapsto$ Vision, Vision $\mapsto$ Audio, Vision $\mapsto$ Text). 
The first subtable summarizes the average accuracy across all tasks for each modality configuration, while the remaining subtables present detailed performance within each task family. 
% Scores are color-coded for readability: \colorbox{lightred}{$<30$}, \colorbox{lightorange}{30–60}, \colorbox{lightgreen}{60–90}, and \colorbox{lightblue}{$\geq90$}, with the best value in each column highlighted in \textbf{bold}. 
The highest scores are \textbf{bolded}, and the second highest are \underline{underlined}.
For each model, we also report the overall average accuracy (\emph{Avg.}) and standard deviation (\emph{Std.}) across the six configurations to quantify robustness to modality shifts. 
Details of the human evaluation are provided in Appendix~\ref{app:human}.

% In the top block of the table, we further report the average accuracy of each modality configuration across all five tasks, providing a comprehensive comparison of model behavior under different input–output settings.

% \textbf{Overall performance across models.} 
% For each task family, we also compute the \emph{average} across all six modality configurations, as well as the \emph{standard deviation (Std.)}, which reflects robustness under modality shifts.  
% Frontier proprietary models such as Gemini consistently outperform open-source alternatives, with Qwen2.5-Omni and EchoInk-R1 representing the strongest open baselines. 
% However, performance gaps remain significant, particularly on the more challenging interleaved settings. 

% \textbf{Performance by task.} 
% Breaking down by task families reveals clear asymmetries: temporal and spatial reasoning tasks are the most difficult across the board, while perception and linguistic tasks are relatively stronger. 
% External knowledge retrieval falls in between, reflecting mixed reliance on modality grounding and language priors. 
% When comparing the best open-source models (Qwen2.5-Omni and EchoInk-R1) with the frontier closed-source Gemini 2.5 Pro, we find that performance is relatively close on perceptual and linguistic tasks, but large gaps remain in spatial and temporal reasoning, highlighting these domains as the primary bottlenecks.
\vspace{-0.2em}
\textbf{Performance by task families.}
% Overall, the Gemini 2.0 and 2.5 series outperform all open-source systems. Among open-source models, Qwen2.5-Omni and EchoInk-R1 are the strongest baselines, surpassing Gemini 1.5 Pro by 3.6 and 4.2 points, respectively.
% Across the five task families, spatial and temporal reasoning remain the most challenging (Gemini 2.5 Pro achieves 50.1 and 60.8), whereas perception and linguistic tasks show relatively higher performance (75.9 and 76.8).
% The performance divergence between open- and closed-source models emerges not only in spatial and temporal reasoning but also in external knowledge tasks — while Qwen2.5-Omni and EchoInk-R1 reach comparable scores to Gemini 2.5 Pro on perception, the latter attains a markedly higher 89.3 on external knowledge.
% These results highlight key bottlenecks for open-source models: closed-source systems likely benefit from broader web-scale pretraining data and exhibit stronger spatial-temporal reasoning capabilities
Overall, the Gemini~2.0 and~2.5 series outperform all open-source models. 
Among open models, Qwen2.5-Omni and EchoInk-R1 are the strongest baselines, surpassing Gemini~1.5~Pro by 3.6 and~4.2 points, respectively. 
Across the five task families, spatial and temporal reasoning remain the most challenging (Gemini~2.5~Pro achieves 50.1 and~60.8), whereas perception and linguistic tasks reach higher accuracy (75.9 and~76.8). 
The performance gap between open- and closed-source models extends beyond spatial and temporal reasoning to external knowledge: while Qwen2.5-Omni and EchoInk-R1 perform comparably to Gemini~2.5~Pro on perception, the latter attains 89.3 on external knowledge. 
These results highlight persistent bottlenecks in open-source models, as closed-source models likely benefit from broader web-scale pretraining and stronger spatial–temporal reasoning capabilities.

% \textbf{Performance by modality configuration.} 
% When grouping results by modality mappings, we observe three distinct trends. 
% (1) First, Vision $\mapsto$ Text consistently outperforms Audio $\mapsto$ Text, showing that models ground visual content more reliably than auditory input. 
% (2) Second, Audio $\mapsto$ Vision and Vision $\mapsto$ Audio settings without text as a semantic pivot—achieve the lowest accuracies, underscoring the difficulty of anchor-free cross-modal retrieval. 
% (3) Third, different models show distinct modality patterns, indicating divergent training emphases or pretraining corpora. 

% To quantify \emph{modality shift robustness}, we further report the standard deviation of accuracy across the six directional mappings: lower variance indicates that the model maintains more stable performance under modality shifts. 
% Among these models, open-source systems such as Qwen2.5-Omni (Std.\ 10.1) and EchoInk-R1 (Std.\ 11.3) show relatively stable behavior compared to other open baselines. 
% Gemini 2.5 Pro achieves the strongest overall accuracy (70.6) with low variance (Std.\ 11.7), showing both capability and stability under modality shifts. However, the task-level results reveal weaknesses especially in spatial reasoning, indicating that high overall accuracy does not guarantee balanced competence across domains.
% Other models like Gemini 1.5 Pro and Baichuan Omni 1.5 display larger fluctuations (Std.\ $>$14), revealing weaker robustness to modality changes despite competitive averages.
% \textbf{Performance by modality configuration.}
\textbf{Performance by modality configurations.}
We also analyze performance consistency across modality configurations on the same tasks and observe clear divergences. 
Vision–text settings consistently outperform audio–text ones, confirming that visual representations are more strongly grounded than audio. 
In perception tasks, accuracy exceeds 90\% with vision–text inputs but drops by over 20 points with audio–text. 
Audio–vision combinations without textual anchors yield the lowest scores, highlighting the difficulty of aligning heterogeneous signals. 
Among SOTA models, Gemini~2.5~Pro (Avg.\ 70.6, Std.\ 11.7) shows the best balance of accuracy and stability, 
while Qwen2.5-Omni (Std.\ 10.1) and EchoInk-R1 (Std.\ 11.3) are the most consistent open models. 
By contrast, Gemini~1.5~Pro and Baichuan~Omni~1.5 have standard deviations exceeding 14, reflecting weaker robustness to modality variation.

% \subsection{Modality Disparity Analysis}
% A central question in omni-language modeling is whether systems exhibit balanced understanding of different sensory modalities. Since XModBench systematically composes tasks across vision, audio, and text, it enables us to directly compare how well models ground visual versus auditory inputs within each task domain.
% To quantify disparity, we fix the semantic content and compare accuracy under different modality settings. 
% For example, between text-to-audio and text-to-vision:
% \[
% \Delta_{T} = Acc(T \mapsto A) - Acc(T \mapsto V),
% \]
% and between audio-to-text and vision-to-text:
% \[
% \Delta_{C} = Acc(A \mapsto T) - Acc(V \mapsto T).
% \]
% Larger absolute values of $\Delta$ indicate stronger modality dependence.

% \subsection{Modality Disparity Analysis}
% \subsection{Modality Disparity Analysis}  
\vspace{-1em}
\subsection{Modality Disparity Analysis}
\vspace{-0.5em}
\label{sec:exp_modality}
% “\textit{Audio remains the most challenging modality.}”  
% \begin{quote}
%  \centering
% \textit{Audio remains the most challenging modality.}
% % , with vision showing moderate gaps and text the most reliable.
% \end{quote}

% \begin{takeawaybox}
% \textbf{Takeaway.} Audio remains the most challenging modality.
% \end{takeawaybox}

% A key challenge for omni-language models is whether they handle audio, vision or text with equal capability rather than favoring one modality. 
% XModBench enables this comparison by instantiating the same semantic content under different modality settings. We quantify disparity by contrasting paired configurations, e.g., $\Delta_{T\text{ vs.\ }V} = \left(Acc_{A \mapsto \textbf{V}} - Acc_{A \mapsto \textbf{T}}\right)+  \left(Acc_{\textbf{V} \mapsto {A}}  - Acc_{\textbf{T} \mapsto A} \right)$ that design compares combinations that differ only by substituting text with vision, thereby controlling for task semantics and isolating the effect of modality substitution on performance.
% Results in Fig.~\ref{fig:exp_cap} reveal a consistent trend: \textbf{audio–text} disparity is the largest (e.g., $-49$ for Gemini~2.5~Pro), reflecting weaker grounding of audio domain against text; while \textbf{audio–vision} disparity is more pronounced (about $-33$) showing the difficulty of directly aligning heterogeneous signals. \textbf{Vision–text} disparity is relatively small (around $-15$). 

A key challenge for OLLMs is whether they handle audio, vision, and text equally rather than favoring one modality. 
\textsc{XModBench} enables this by instantiating identical semantics across modality settings. The disparity is defined as  
$\Delta_{T\text{ vs.\ }V} = (Acc_{A \mapsto \mathbf{V}} - Acc_{A \mapsto \mathbf{T}}) + (Acc_{\mathbf{V} \mapsto A} - Acc_{\mathbf{T} \mapsto A})$,
\begin{wrapfigure}{r}{0.55\textwidth}
  \centering
  \vspace{-10pt}
  \includegraphics[width=\linewidth]{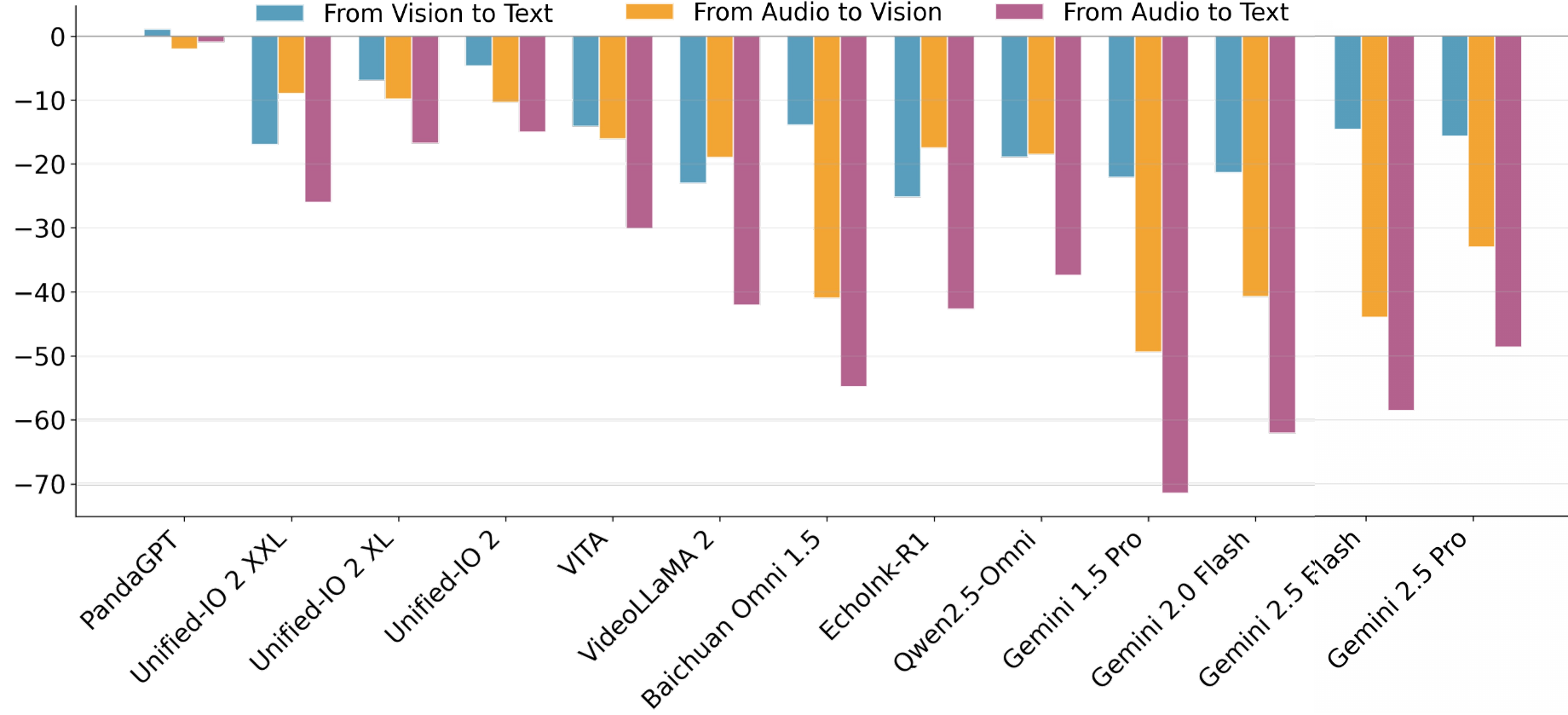}
    \caption{Modality disparity across different configurations. Negative scores indicate performance gaps, with the largest disparities observed between audio and text.\vspace{-2em}}
  \label{fig:exp_cap}
\end{wrapfigure}
which quantify the paired subtraction, e.g.,  compares configurations that differ only by substituting \textbf{text} with \textbf{vision}, thereby isolating the effect of modality substitution on accuracy. 
Results in Fig.~\ref{fig:exp_cap} show that $\Delta_{T\text{ vs.\ }A}$ exhibits the strongest disparity ($\!-49$ for Gemini~2.5~Pro), $\Delta_{V\text{ vs.\ }A}$ is moderate ($\!-33$), and $\Delta_{T\text{ vs.\ }V}$ remains smallest ($\!-15$). 
\textit{These findings highlight audio as the most challenging modality, with vision showing moderate gaps and text remaining the most robust.}

% \vspace{-0.5em}
% \begin{wrapfigure}{r}{0.55\textwidth}
%   \centering
%   \vspace{-20pt}
%   \includegraphics[width=\linewidth]{figures/exp_cap.pdf}
%     \caption{Modality disparity across different configurations. Negative scores indicate performance gaps, with the largest disparities observed between audio and text.\vspace{-5pt}}
%   \label{fig:exp_cap}
% \end{wrapfigure}
% \vspace{-1em}

% We analysis the Modality understanding ablity. As we donot contain pure text understanding, we especially start from basic task for vision and audio understanding..
% We begin by examining modality-level performance, focusing on the fundamental grounding of sensory input into text. A clear asymmetry emerges: \textbf{Vision $\mapsto$ Text} consistently outperforms Audio $\mapsto$ Text, highlighting that visual grounding is far more mature, while audio remains a major bottleneck across models.

% \subsection{Modality Interleaving Analysis}
% \subsection{Modality Balance}
% Beyond understanding inputs, an equally important question is how well models can \emph{generate or align outputs} across modalities. By examining tasks where text, vision, or audio serve as the candidates, we can assess whether models exhibit balanced capability in producing consistent outputs across different modalities.
\vspace{-0.5em}
\subsection{Directional Imbalance}  
\vspace{-0.5em}
\label{sec:exp_imbalance}

% \begin{quote}
% \centering
% \textit{Directional imbalance is most pronounced in text–vision and audio–text pairs, possibly reflecting training data biases.}
% \end{quote}

% \begin{takeawaybox}
% \textbf{Takeaway.} Directional imbalance is most pronounced in text-vision and audio-text pairs, possibly reflecting training data biases.
% \end{takeawaybox}

\begin{figure}[h]
  \centering
  \includegraphics[width=\linewidth]{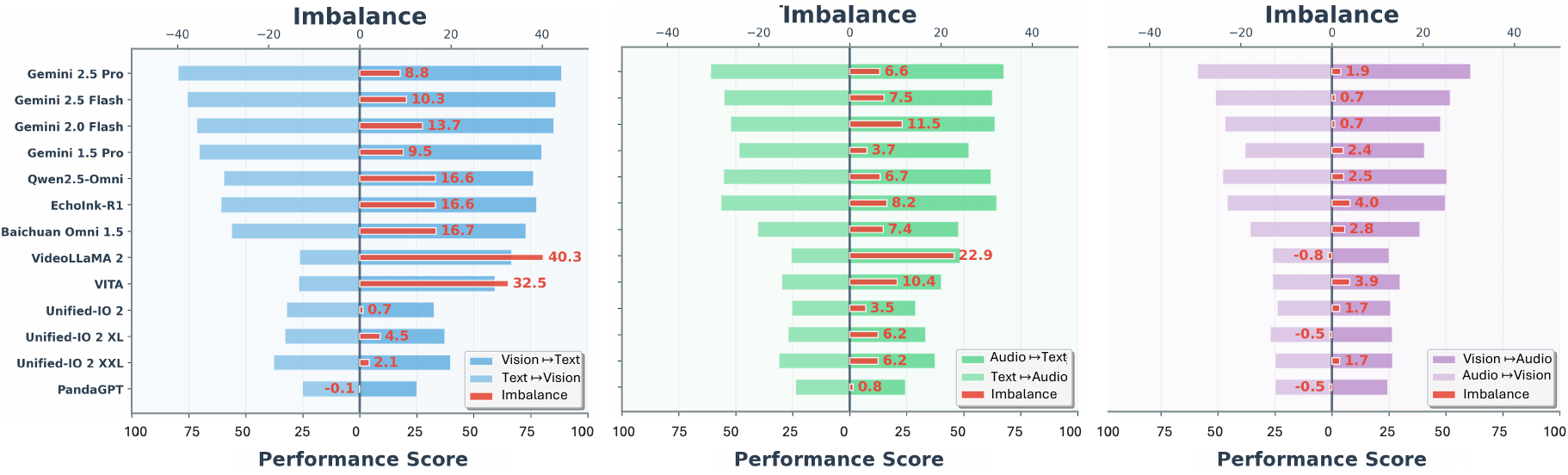}
\caption{Directional imbalance: accuracy gaps between paired inverse settings among audio, vision and text. Models show clear asymmetries, especially in vision–text and audio–text pairs.}
  \label{fig:exp_imbalance}
\end{figure}

We test whether models behave consistently when swapping the roles of context and candidates. 
We define \emph{directional imbalance} as $\Delta_{X \leftrightarrow Y} = Acc(X \mapsto Y) - Acc(Y \mapsto X)$, 
the accuracy gap between inverse configurations for $(X,Y) \in \{(A,T), (V,T), (V,A)\}$. 
As shown in Fig.~\ref{fig:exp_imbalance}, vision–text and audio–text pairs exhibit notable asymmetries: 
Gemini~2.5~Pro drops by $8.8$ points from T$\!\mapsto$V to V$\!\mapsto$T, and Qwen2.5-Omni shows a $16.6$-point gap, 
while audio–text differences remain around $6$–$8$ points. 
By contrast, \textbf{audio–vision} pairs are nearly symmetric but achieve much lower overall accuracy. 
\textit{These findings suggest that directional imbalance mainly arises in text–vision and audio–text pairs, likely reflecting training data biases toward text as the dominant output modality.}
  
% \subsection{Directional Imbalance}
% Beyond overall capability differences, we analyze whether models behave consistently across inverse modality directions and interleaved settings. 
% Text $\mapsto$ Vision versus Vision $\mapsto$ Text and Text $\mapsto$ Audio versus Audio $\mapsto$ Text allow us to measure \emph{directional consistency}, i.e., whether models exhibit balanced performance when text is shifted from context to candidate space. 
% Meanwhile, Audio $\mapsto$ Vision and Vision $\mapsto$ Audio test anchor-free retrieval between non-text modalities, where performance drops most sharply. 
% Together, these analyses expose systematic imbalances in cross-modal grounding, reinforcing the need for benchmarks like XModBench to uncover hidden weaknesses beyond overall accuracy.

\vspace{-1em}

\vspace{-1em}
\subsection{Failure Case Analysis}
\begin{figure}[h]
  \centering
  \includegraphics[width=0.8\linewidth]{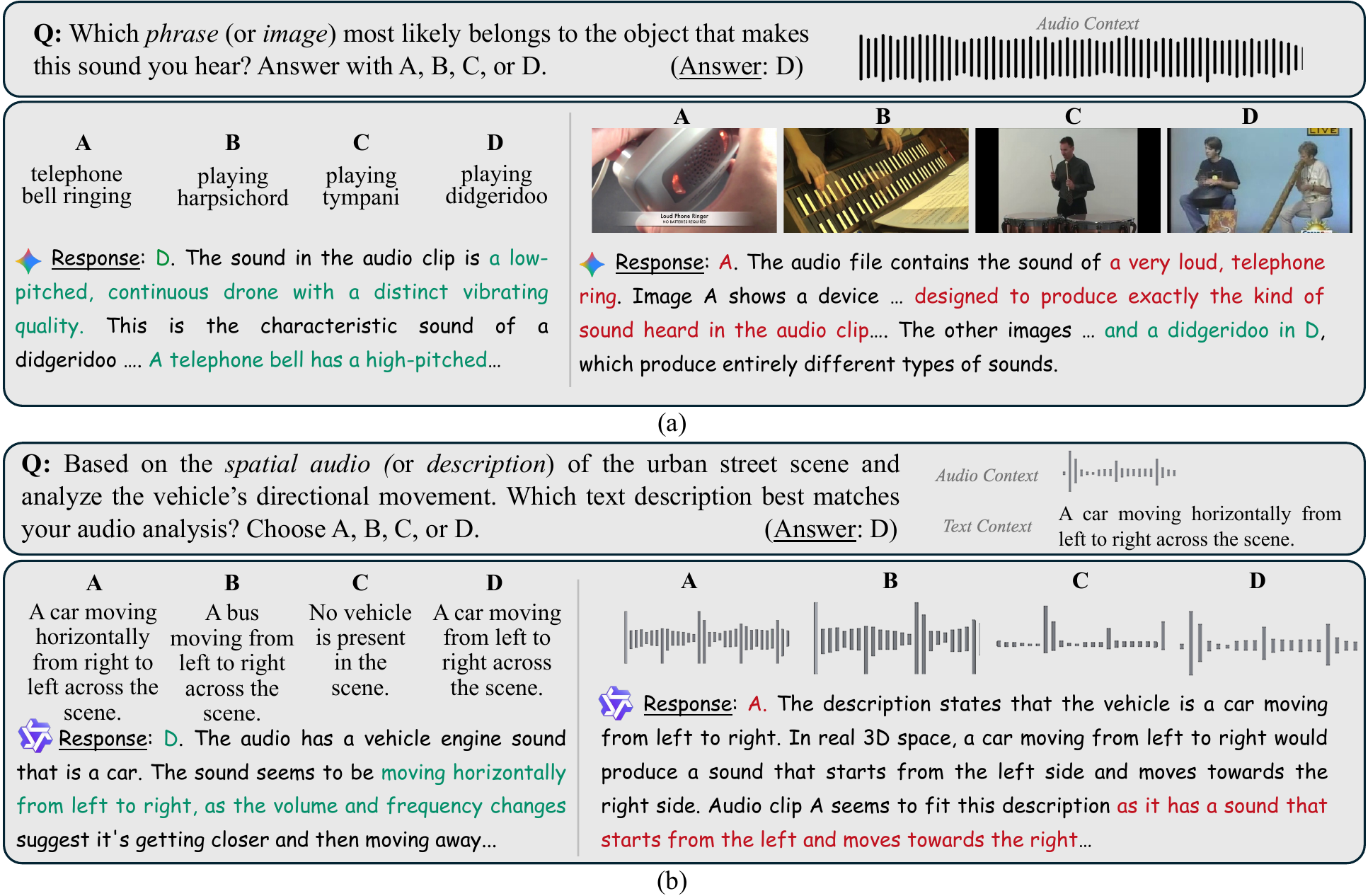}
  \caption{Failure cases. (a)  Gemini 2.5 pro correctly identifies a didgeridoo in text but fails to match it with the corresponding image candidates. (b) shows Qwen2.5-Omni misinterprets spatial motion when switching candidates from text to audio. This cases illustrate asymmetries in cross-modal reasoning.}
\vspace{-1em}
  \label{fig:failure_cases}
\end{figure}

\vspace{-1em}
To better understand model errors, we prompt models like Gemini 2.5 Pro and Qwen2.5-Omni to generate reasoning alongside their answers. As shown in Fig.~\ref{fig:failure_cases}, we observe common failure cases that reflect modality performance gaps and alignment issues. Example (a) shows a mismatch between audio-to-text and audio-to-image reasoning: while the model correctly identifies a didgeridoo by text, it fails to select the matching image, revealing inconsistent grounding. In example (b), Qwen-2.5 Omni misinterprets spatial audio motion when switching from audio-to-text to text-to-audio, reversing the vehicle’s direction. These errors highlight persistent asymmetries in cross-modal reasoning that only emerge when the same task is posed across different modality combinations.

\vspace{-1em}
\subsection{Triple-Domain Question Answering}
\vspace{-0.8em}
\revise{Real-world omni-modal scenarios often present information jointly across modalities rather than in isolation.  
To approximate this setting, we extend \textsc{XModBench} tasks to an audio–visual context, where both sound and vision are provided in the question stem, while the candidates remain in text.}  

We evaluate this dual-context configuration using the Gemini series.  
Compared with single-modality baselines, the results (see Appendix~\ref{app:dual_context}) show modest but consistent gains, indicating that models can benefit from simultaneous multimodal cues.  
However, the improvements are not always additive, suggesting that current models do not yet fully exploit complementary signals across modalities.

\begin{table}[!h]
\centering
\small
% \caption{Overall performance of Gemini models under the dual-context setting 
% (audio+visual context $\mapsto$ text). We compare with pairwise baselines (A $\mapsto$ T and V $\mapsto$ T), 
% and report the stronger unimodal baseline $\max(\text{A $\mapsto$ T}, \text{V $\mapsto$ T})$.
\caption{Performance of Gemini models  using both audio and visual context ($A+V$). Comparing $A+V$ with the best single-modality result ($\max(A, V)$) reveals whether adding more context leads to better reasoning or causes harmful interference.}
\small
\label{tab:dual_context}
\begin{tabular}{l|c|c|c}
\toprule
\textbf{Setting} & \textbf{Gemini 1.5 Pro} & \textbf{Gemini 2.0 Flash} & \textbf{Gemini 2.5 Pro}  \\
\midrule
~~~~~A $\mapsto$ T   & 52.76 & 63.71 & 70.99 \\
~~~~~V $\mapsto$ T   & 79.92 & 85.20 & 88.60 \\\midrule
% A+V $\mapsto$ T & 82.53 (\textbf{+2.61}) & 79.84 (\textbf{-5.36})  &  89.76 (\textbf{+1.16}) \\
A+V $\mapsto$ T & 82.53 (\textbf{\textcolor{ForestGreen}{+2.61}}) & 79.84 (\textbf{\textcolor{Maroon}{-5.36}}) & 89.76 (\textbf{\textcolor{ForestGreen}{+1.16}}) \\
\bottomrule
\end{tabular}
\end{table}

\vspace{-0.5em}
\section{\revise{Key Takeaways for OLLM Development}}
\vspace{-0.8em}
\revise{Our benchmark results serve as a diagnostic tool, revealing how underlying \textbf{data composition} and \textbf{training methodologies} shape OLLMs behaviors. By correlating performance patterns in \ref{sec:exp_modality} and \ref{sec:exp_imbalance} with model architectures and training reports, we derive three critical insights regarding \textit{interleaved data}, \textit{task coverage},  and how these effect OLLMs after \textit{post-training}.}

\vspace{-0.5 em}
\subsection{\revise{Interleaved Data Reduces Directional Imbalance}}\vspace{-1em}
\revise{A key observation from our imbalance analysis is the connection between interleaved training data and the reduction of directional imbalance when modalities are swapped, as seen in the results from Section~\ref{sec:exp_imbalance}.}

Public official reports indicate that models such as \emph{Qwen2.5-Omni}~\citep{xu2025qwen2} and \emph{Gemini-2.5} series~\citep{comanici2025gemini} incorporate massive scale interleaved multimodal corpora (e.g., mixed audio-vision conversations). Our benchmark corroborates this: these models exhibit relatively small performance gaps between Audio$\to$Vision and Vision$\to$Audio tasks. This suggests that encountering modalities interchangeably in context allows the model to build symmetric cross-modal bridges.

Conversely, models trained primarily on non-interleaved datasets exhibit significant directional asymmetry. For instance, in the case of VideoLLaMA~\citep{cheng2024videollama}, despite having strong backbones, models relying on open-source data with limited interleaved audio-vision instruction pairs show a distinct bias. Models fail to generalize when the modalities are swapped (e.g., needing to distinguish between four video or audio clips), indicating that insufficient interleaved supervision hinders directional robustness.
% \end{itemize}

\subsection{Task Coverage Gaps}

With its broad domain coverage, \textbf{XModBench} serves as a diagnostic tool to reveal blind spots in training for both open- and closed-source models. It identifies these gaps through performance inconsistencies, particularly regarding the diversity of audio types.
\vspace{-0.5em}
\begin{itemize}
\item \textbf{Spoken vs.\ Non-Spoken Bias:} Many OLLMs prioritize speech as the primary modality for audio data. Models relying on pre-trained, speech encoders (e.g., Whisper~\citep{radford2023robust} in the \textit{Baichuan}~\citep{li2025baichuan} model) and trained predominantly on spoken language tasks exhibit a sharp performance decline in non-spoken audio domains, such as environmental sound recognition and spatial reasoning. This indicates a significant semantic gap between linguistic speech and general acoustic signals, resulting in a weakness for comprehensive audio understanding.
% \end{itemize}

\item \revise{\textbf{Deficiencies in Music Understanding and Spatial Reasoning:} Distinct gaps in specific categories shows evidence of missing training data. For example, despite its high overall capacity, \emph{Gemini~1.5} demonstrates limited capability in musical reasoning, suggesting an absence of related data in its training corpus. Similarly, \emph{EchoInk-R1} struggles with spatial-vision tasks relative to related families, pointing to a lack of spatial-centric visual content.}
\end{itemize}

\subsection{\revise{The Impact of Post-Training on Alignment}}

A comparative analysis of \emph{EchoInk-R1} and the \emph{Qwen2.5-Omni} series illustrates how post-training strategies can reshape and even degrade multimodal alignment when data is weakness, despite the advance reinforcement learning used to enhances cross-modal reasoning
\citep{shao2024deepseekmath}. \emph{Qwen2.5-Omni} integrates interleaved multimodal conversations during post-training, \emph{EchoInk-R1} is further fine-tuned on the data derived from OmniInstruct corpus~\citep{li2024omnibench} focus on spoken instruction following. This divergence in training objectives leads to notable behavioral differences:
\begin{itemize}
% \item \revise{\textbf{Alignment Erosion:} \textit{EchoInk-R1 }shows decreased performance in cross-modal (AV/VA) tasks compared to \textit{Qwen2.5-Omni}. This suggests that aggressive fine-tuning on spoken-only instructions may cause catastrophic forgetting of the fine-grained cross-modal grounding acquired during pre-training.}

% \item  While \textit{EchoInk-R1} is built upon the well-aligned \textit{Qwen2.5-Omni} base, it exhibits a noticeable performance decline in cross-modal (AV/VA) tasks. This suggests that the balanced cross-modal bridges built through interleaved pre-training can be eroded if the post-training stage shifts to modality-specific (i.e., spoken-only) instruction tuning
\vspace{-0.5em}
\item Although \textit{EchoInk-R1} is based on the well-aligned \textit{Qwen2.5-Omni}, it shows \textbf{larger modality disparities and directional imbalances}.  Fig.\ref{fig:exp_cap} shows it performs worse in Audio$\to$Text and Vision$\to$Text modality disparity, and exhibits higher imbalance in Audio-Vision and Audio-Text pairs in Fig.~\ref{fig:exp_imbalance}. This suggests that the cross-modal alignment from interleaved pre-training can be eroded if the post-training stage shifts to non-interleaved data.

\item Despite heavy tuning, \textit{EchoInk} fails to improve in spatial tasks (Fig.~\ref{tab:multimodal_performance_lite}), showing that post-training requires \textbf{comprehensive task coverage}. Incomplete fine-tuning cannot fill these gaps and may even degrade the model's multimodal capabilities.
\end{itemize}

Beyond simple evaluation, XModBench serves as a diagnostic tool for model development. While training data for SOTA models remains a black box, XModBench reveals the impact of training strategies and task coverage through controlled comparisons from them. Our open-source generation tools also offer the potential to extend to new domains. By using a unique modality-swap design, XModBench exposes hidden flaws that current benchmarks miss, making it essential for building robust next generation omni-modal language models.

% \section{Conclusion}

% We present XModBench, a comprehensive benchmark for diagnosing cross-modal consistency in multimodal large language models. By systematically interleaving audio, vision, and text across diverse tasks, XModBench enables fine-grained analysis of modality understanding, robustness to modality shifts, and directional asymmetries. Our results reveal that while current omni-models show promising performance on individual modalities, they often struggle with consistent reasoning across modality changes. XModBench offers a rigorous testbed to guide future improvements in truly modality-agnostic AI systems.
% % opensource?
\vspace{-0.5em}

\section{Conclusion}
\vspace{-0.5em}
We introduced \textbf{XModBench}, a benchmark for diagnosing cross-modal consistency in omni-language models.  
By systematically interleaving audio, vision, and text across diverse tasks, XModBench enables fine-grained evaluation of modality disparities, directional imbalances, and modality invariant capability.  
Our results show that audio remains the most challenging modality, that models often behave asymmetrically in inverse settings such as text-vision and audio-text, and that combining audio and vision yields only modest gains.  
Overall, while current systems are strong in perception and language, they still lack stable and consistent reasoning across modalities, leaving ample room for progress toward truly modality-agnostic intelligence.  

\newpage

\section*{Ethics statement}

Our study does not involve private or sensitive personal data. All audiovisual samples are obtained from publicly available official sources, including previously published research datasets, content hosted on established open source platforms such as Hugging Face and Kaggle.

For all newly generated labels and annotations, we perform manual verification to ensure correctness and to remove any potentially inappropriate content. All web-curated data are from publicly accessible and previously published sources without requiring special authentication. All materials are used solely for non-commercial academic research. We do not redistribute copyrighted video or audio; only derived features, annotations, and evaluation results are released.

\bibliography{iclr2026_conference}
\bibliographystyle{iclr2026_conference}

\newpage
\appendix

\section*{Supplementary Material}
\section*{Table of Content}
\begin{description}
  \item[A] Mini Benchmark Result \dotfill \pageref{sec:lite}
  \item[B] Modality Configuration Score under Five Tasks \dotfill \pageref{sec:modality}
  \item[C] Task-Specified Model Performance \dotfill \pageref{sec:taskperf}
  \begin{description}
    \item[\hspace{1em}C.1] Task 1: Perceptual Task \dotfill \pageref{sec:t1}
    \item[\hspace{1em}C.2] Task 2: Spatial Reasoning \dotfill \pageref{sec:t2}
    \item[\hspace{1em}C.3] Task 3: Temporal Reasoning \dotfill \pageref{sec:t3}
    \item[\hspace{1em}C.4] Task 4: Linguistic Task \dotfill \pageref{sec:t4}
    \item[\hspace{1em}C.5] Task 5: External Knowledge \dotfill \pageref{sec:t5}
    
  \end{description}
  \item[D] Evaluation Cost \dotfill \pageref{sec:cost}
  \item[E] Interleaving Visual Audio Input \dotfill \pageref{app:dual_context}
  \item[F] Human Survey \dotfill \pageref{app:human}
  \item[G] Technical Details in Triplet Data Collection \dotfill \pageref{app:data}
  \begin{description}
    \item[\hspace{1em}G.1] Perceptual Recognition \dotfill \pageref{sec:data:perc}
    \item[\hspace{1em}G.2] Spatial Reasoning \dotfill \pageref{sec:data:spat}
    \item[\hspace{1em}G.3] Temporal Reasoning \dotfill \pageref{sec:data:temp}
    \item[\hspace{1em}G.4] Linguistic Understanding \dotfill \pageref{sec:data:ling}
    \item[\hspace{1em}G.5] External Knowledge \dotfill \pageref{sec:data:knwl}
  \end{description}
  \item[H] LLM Usage \dotfill \pageref{sec:llm}
\end{description}
\newpage

\section{Mini Benchmark result}\label{sec:lite}
\revise{We will release a standardized {6k-sample XModBench-Lite} benchmark, consisting of 
{5 task families} $\times$ {6 modality–configuration settings}, with 
{200 examples per setting}. The dataset is balanced across both task families 
and modality directions. The overall performance (see Tab.~\ref{tab:lite_results}) trends and error patterns closely 
mirror those reported in Tab.\ref{tab:multimodal_performance_lite} of the main paper. }

\begin{table*}[htbp]
  \centering
  \caption{Performance on 6k version of XModBench }
  \label{tab:lite_results}
  \resizebox{\textwidth}{!}{%
\begin{tabular}{l|ccccc|cccccc|c}
      \toprule
      \multirow{2}{*}{\textbf{Model}} & \multicolumn{5}{c|}{\textbf{Accuracy on 5 Task Families}} & \multicolumn{6}{c|}{\textbf{Modality Configuration}} & \multirow{2}{*}{\textbf{Avg.}} \\
      \cmidrule(lr){2-6} \cmidrule(lr){7-12}
       & \textbf{Perc.} & \textbf{Spat.} & \textbf{Temp.} & \textbf{Ling.} & \textbf{Knwl.} & \textbf{A $\to$ T} & \textbf{A $\to$ V} & \textbf{T $\to$ A} & \textbf{T $\to$ V} & \textbf{V $\to$ A} & \textbf{V $\to$ T} & \\
      \midrule
      
      % Baseline
      w/o context & 25.3 & 25.1 & 24.8 & 24.4 & 25.2 & 26.5 & 24.8 & 24.2 & 24.1 & 25.5 & 25.1 & 25.0 \\
      \midrule

      % Red Highlighted VL Models
      {Qwen2.5-VL} & {91.5} & {51.9} & 40.5 & {84.3} & 76.5 & - & - & - & 68.2 & - & 72.8 & 60.5 \\
      {Intern3.5-VL} & {88.2} & 41.8 & {48.5} & 75.8 & 62.4 & - & - & - & 46.5 & - & 73.1 & {69.8} \\
      \midrule

      % Other Models
      PandaGPT & 24.9 & 25.3 & 23.8 & 24.7 & 21.3 & 25.2 & 25.5 & 22.8 & 24.9 & 24.8 & 23.1 & 24.4 \\
      Unified-IO 2 & 36.5 & 24.8 & 24.1 & 31.2 & 27.5 & 29.8 & 24.2 & 25.5 & 32.1 & 25.9 & 33.5 & 28.4 \\ 
      Unified-IO 2 XL & 42.5 & 26.3 & 26.0 & 32.5 & 30.6 & 33.8 & 26.8 & 27.5 & 34.2 & 27.1 & 38.8 & 31.2 \\ 

      Unified-IO 2 XXL & 44.1 & 29.0 & 27.3 & 32.9 & 34.7 & 38.0 & 26.3 & 31.8 & 38.5 & 27.3 & 40.5 & 33.6 \\ 

      VideoLLaMA 2 & 46.1 & 34.2 & 29.0 & 37.5 & 37.4 & 49.1 & 26.2 & 26.0 & 27.3 & 25.5 & 67.8 & 36.9 \\

      VITA & 35.6 & 31.8 & 29.2 & 46.1 & 30.5 & 45.2 & 26.5 & 26.8 & 26.1 & 24.5 & 52.5 & 33.6 \\
      Baichuan Omni 1.5 & 59.5 & 35.5 & 30.8 & 63.9 & 57.2 & 48.5 & 36.2 & 41.1 & 57.2 & 39.1 & 74.5 & 49.5 \\
      EchoInk-R1 & 73.1 & 35.8 & 36.4 & 73.3 & 72.4 & 66.5 & 42.1 & 56.0 & 65.5 & 46.8 & 72.3 & 53.2 \\
      Qwen2.5-Omni & \textbf{78.6} & 37.1 & 31.2 &  74.2 & 77.8 & \underline{69.5} & 45.2 & 54.5 & 58.1 & \underline{56.8} & 74.5 & 51.4 \\
      \midrule

      % Gemini Family
      Gemini 1.5 Pro & 56.8 & 40.8 & 38.0 & 71.0 & 69.9 & 53.1 & 38.5 & 49.2 & 71.2 & 41.4 & 80.4 & 55.7 \\
      Gemini 2.0 Flash & 65.4 & 48.9 & 41.5 & 72.2 & 71.2 & 68.5 & 44.2 & 50.1 & 74.3 & 47.5 & \textbf{88.9} & 67.2 \\
      Gemini 2.5 Flash & 66.5 & \underline{47.1} & \underline{45.3} & \underline{74.4}   & \underline{81.2} & 64.8 & \underline{49.4} & \underline{57.5} & \textbf{77.2} & 51.0 & 81.6 & \underline{68.6} \\
      Gemini 2.5 Pro & \underline{74.8} & \textbf{59.3} & \textbf{60.2} &  \textbf{75.8} & \textbf{89.1} & \textbf{76.8} & \textbf{50.5} & \textbf{63.2} & \underline{75.1} & \textbf{61.2} & \underline{82.5} & \textbf{71.8} \\
      \bottomrule
    \end{tabular}
  }
\end{table*}

\newpage
\section{Modality configuration score under five task}\label{sec:modality}

Table~\ref{tab:multimodal_performance} reports the detailed results for all six
modality–configuration settings (A$\mapsto$T, A$\mapsto$V, T$\mapsto$A, T$\mapsto$V,
V$\mapsto$A, V$\mapsto$T) across the five task families in XModBench
(Perception, Spatial, Temporal, Linguistic, and Knowledge), as well as the
overall average score on the full benchmark. 
% Scores are color-coded for readability: \colorbox{lightred}{$<30$}, \colorbox{lightorange}{30–60}, \colorbox{lightgreen}{60–90}, and \colorbox{lightblue}{$\geq90$}, with the best value in each column highlighted in \textbf{bold}. 
% For each model, we show how
% performance varies when changing the source and target modalities, which allows
% us to disentangle \emph{modality-directional imbalance} from general task
% difficulty. 

\begin{table}[!h]
\centering
\small
% \caption{Performance comparison of multimodal AI models across different tasks showing cross-modal translation performance across audio, text, and vision modalities. \textbf{Note:} $A \mapsto T$ denotes the setting where audio modality is context and choose text canditates.  }
\caption{Results on \textbf{XModBench} across 5 task families and 6 predefined cross-modal directions among \textbf{T}ext, \textbf{V}ision, and \textbf{A}udio. The first block reports the average accuracy across all tasks, followed by Task~1–5 (Perception, Spatial, Temporal, Linguistic, External knowledge). Scores are color-coded as \colorbox{lightred}{$<30$}, \colorbox{lightorange}{30–60}, \colorbox{lightgreen}{60–90}, \colorbox{lightblue}{$\geq 90$}, with the best in each column highlighted in \textbf{bold}.}

% Audio to Text, A $\mapsto$ V: Audio to Vision, TA: Text to Audio, T $\mapsto$ V: Text to Vision, V $\mapsto$ A: Vision to Audio, V $\mapsto$ T: Vision to Text.
\label{tab:multimodal_performance}
\resizebox{\textwidth}{!}{
\begin{tabular}{l|cccccc|c|c|cccccc|c|c}
\toprule
 & \multicolumn{8}{c|}{\textbf{Overall Average}} & \multicolumn{8}{c}{\textbf{Task 1 - Perception}} \\
\cmidrule(lr){2-9} \cmidrule(lr){10-17}
\textbf{Model} & \textbf{A $\mapsto$ T} & \textbf{A  $\mapsto$ V} & \textbf{T $\mapsto$ A} & \textbf{T $\mapsto$ V} & \textbf{V $\mapsto$ A} & \textbf{V $\mapsto$ T} & \textbf{\textit{Avg.}} & \textbf{\textit{Std.}} & \textbf{A $\mapsto$ T} & \textbf{A $\mapsto$ V} & \textbf{T $\mapsto$ A} & \textbf{T $\mapsto$ V} & \textbf{V $\mapsto$ A} & \textbf{V $\mapsto$ T} & \textbf{\textit{Avg.}} & \textbf{\textit{\textit{Std.}}} \\
\midrule
PandaGPT & \perfcell{24.5} & \perfcell{25.0} & \perfcell{23.8} & \perfcell{25.2} & \perfcell{24.5} & \perfcell{25.1} & 24.7 & {0.5} & \perfcell{24.5} & \perfcell{24.7} & \perfcell{24.8} & \perfcell{24.5} & \perfcell{24.6} & \perfcell{24.7} & 24.6 & {0.1} \\
Unified-IO 2 & \perfcell{28.9} & \perfcell{24.0} & \perfcell{25.4} & \perfcell{32.0} & \perfcell{25.7} & \perfcell{32.7} & 28.1 & {3.7} & \perfcell{35.5} & \perfcell{25.3} & \perfcell{26.3} & \perfcell{55.9} & \perfcell{29.1} & \perfcell{44.7} & 36.1 & {12.1} \\
Unified-IO 2 XL & \perfcell{33.3} & \perfcell{27.0} & \perfcell{27.1} & \perfcell{32.9} & \perfcell{26.5} & \perfcell{37.4} & 30.7 & {4.5} & \perfcell{53.3} & \perfcell{27.9} & \perfcell{30.3} & \textbf{\perfcell{59.1}} & \perfcell{27.6} & \perfcell{55.0} & 42.2 & {15.0} \\
Unified-IO 2 XXL & \perfcell{37.4} & \perfcell{25.0} & \perfcell{31.2} & \perfcell{37.8} & \perfcell{26.7} & \perfcell{39.9} & 33.0 & {6.3} & \perfcell{55.0} & \perfcell{26.9} & \perfcell{39.0} & \perfcell{64.2} & \perfcell{26.7} & \perfcell{50.2} & 43.7 & {15.4} \\
VideoLLaMA 2 & \perfcell{48.6} & \perfcell{26.0} & \perfcell{25.7} & \perfcell{26.5} & \perfcell{25.2} & \perfcell{66.8} & 36.5 & {17.4} & \textbf{\perfcell{74.7}} & \perfcell{26.6} & \perfcell{28.3} & \perfcell{26.8} & \perfcell{26.5} & \perfcell{91.5} & 45.7 & {29.4} \\

VITA & \perfcell{40.2} & \perfcell{26.0} & \perfcell{29.8} & \perfcell{26.8} & \perfcell{29.9} & \perfcell{59.3} & 35.4 & {12.8} & \perfcell{37.1} & \perfcell{25.4} & \perfcell{27.0} & \perfcell{23.7} & \perfcell{26.4} & \perfcell{69.1} & 34.8 & {17.5} \\
Baichuan Omni 1.5 & \perfcell{47.8} & {\perfcell{35.8}} & \perfcell{40.5} & \perfcell{56.2} & \perfcell{38.6} & \perfcell{73.0} & 48.7 & {14.0} & \perfcell{42.7} & \perfcell{36.3} & \perfcell{45.6} & \perfcell{87.8} & \perfcell{50.3} & \perfcell{90.7} & 58.9 & {24.0} \\
EchoInk-R1 & \perfcell{64.6} & \perfcell{45.9} & \perfcell{56.4} & \perfcell{60.9} & \perfcell{49.9} & \perfcell{77.6} & 59.2 & {11.3} & \perfcell{74.1} & \perfcell{58.5} & \textbf{\perfcell{69.3}} & \perfcell{91.6} & \perfcell{67.7} & \perfcell{93.4} & 75.8 & {13.9} \\
Qwen2.5-Omni & \perfcell{62.0} & \perfcell{48.0} & \perfcell{55.4} & \perfcell{59.6} & {\perfcell{50.5}} & \perfcell{76.3} & 58.6 & {10.1} & \perfcell{72.9} & {\perfcell{59.1}} & \perfcell{69.2} & \perfcell{91.2} & \perfcell{68.5} & \perfcell{92.0} & 75.5 & {13.3} \\  \midrule
Gemini 1.5 Pro & \perfcell{52.4} & \perfcell{38.2} & \perfcell{48.6} & \perfcell{70.4} & \perfcell{40.7} & \perfcell{79.9} & 55.0 & {16.7} & \perfcell{41.0} & \perfcell{27.9} & \perfcell{45.0} & \perfcell{95.8} & \perfcell{32.1} & \perfcell{95.3} & 56.2 & {31.1} \\
Gemini 2.0 Flash & \perfcell{63.7} & \perfcell{49.0} & \perfcell{52.2} & \perfcell{71.5} & \perfcell{47.6} & \perfcell{85.2} & 61.2 & {15.2} & \perfcell{56.8} & \perfcell{45.0} & \perfcell{54.2} & \perfcell{92.7} & \perfcell{55.1} & \perfcell{93.4} & 66.2 & {21.2} \\
Gemini 2.5 Flash & \perfcell{62.6} & \perfcell{51.2} & \perfcell{55.1} & \perfcell{75.7} & \perfcell{51.9} & \perfcell{86.0} & 63.7 & {14.2} & \perfcell{52.6} & \perfcell{44.3} & \perfcell{53.4} & \perfcell{95.4} & \perfcell{56.0} & \perfcell{95.0} & 66.1 & {22.8} \\
Gemini 2.5 Pro & \textbf{\perfcell{71.0}} & \textbf{\perfcell{58.9}} & \textbf{\perfcell{64.4}} & \textbf{\perfcell{79.8}} & \textbf{\perfcell{60.8}} & \textbf{\perfcell{88.6}} & \textbf{70.6} & {11.7} & \perfcell{62.3} & \perfcell{57.4} & \perfcell{68.5} & \textbf{\perfcell{97.1}} & \textbf{\perfcell{72.6}} & \textbf{\perfcell{97.6}} & \textbf{75.9} & {17.4} \\
\midrule
Human & {92.4} & 91.5 & 91.1 & 91.8 & 86.4 & 95.6 & 91.5 & 3.0 & 92.9 & 94.2 & 91.3 & 89.2 & 85.4 & 92.9 & 91.0 & 3.2 \\

\bottomrule
\end{tabular}
}

\resizebox{\textwidth}{!}{
\begin{tabular}{l|cccccc|c|c|cccccc|c|c}
\toprule
 & \multicolumn{8}{c|}{\textbf{Task 2 - Spatial}} & \multicolumn{8}{c}{\textbf{Task 3 - Temporal}} \\
\cmidrule(lr){2-9} \cmidrule(lr){10-17}
\textbf{Model} & \textbf{A $\mapsto$ T} & \textbf{A $\mapsto$ V} & \textbf{T $\mapsto$ A} & \textbf{T $\mapsto$ V} & \textbf{V $\mapsto$ A} & \textbf{V $\mapsto$ T} & \textbf{\textit{Avg.}} & \textbf{\textit{Std.}} & \textbf{A $\mapsto$ T} & \textbf{A $\mapsto$ V} & \textbf{T $\mapsto$ A} & \textbf{T $\mapsto$ V} & \textbf{V $\mapsto$ A} & \textbf{V $\mapsto$ T} & \textbf{\textit{Avg.}} & \textbf{\textit{Std.}} \\
\midrule
PandaGPT & \perfcell{25.5} & \perfcell{26.6} & \perfcell{26.0} & \perfcell{27.2} & \perfcell{25.8} & \perfcell{23.1} & 25.7 & {1.4} & \perfcell{21.9} & \perfcell{25.3} & \perfcell{24.8} & \perfcell{26.0} & \perfcell{24.5} & \perfcell{23.9} & 24.4 & {1.4} \\
Unified-IO 2 & \perfcell{26.0} & \perfcell{20.7} & \perfcell{22.4} & \perfcell{25.0} & \perfcell{23.1} & \perfcell{24.7} & 23.6 & {1.9} & \perfcell{22.7} & \perfcell{22.4} & \perfcell{25.1} & \perfcell{24.3} & \perfcell{25.8} & \perfcell{22.4} & 23.8 & {1.5} \\ 
Unified-IO 2 XL & \perfcell{24.8} & \perfcell{23.0} & \perfcell{25.8} & \perfcell{26.0} & \perfcell{26.0} & \perfcell{24.5} & 25.0 & {1.2} & \perfcell{22.3} & \perfcell{24.5} & \perfcell{28.8} & \perfcell{22.1} & \perfcell{26.0} & \perfcell{32.7} & 26.1 & {4.1} \\
Unified-IO 2 XXL & \perfcell{29.6} & \perfcell{23.6} & {\perfcell{30.9}} & \perfcell{25.5} & \perfcell{29.5} & \perfcell{30.7} & 28.3 & {3.0} & \perfcell{24.3} & \perfcell{27.4} & \perfcell{25.3} & \perfcell{29.6} & \perfcell{25.2} & \perfcell{34.4} & 27.7 & {3.8} \\
% \midrule
VideoLLaMA 2 & {\perfcell{43.9}} & \perfcell{27.8} & \perfcell{24.4} & \perfcell{27.5} & \perfcell{25.2} & \perfcell{54.3} & 33.9 & 12.3 & \perfcell{31.0} & \perfcell{25.0} & \perfcell{27.7} & \perfcell{25.9} & \perfcell{25.8} & \perfcell{39.8} & 29.2 & {5.6} \\
VITA & \perfcell{42.3} & \perfcell{28.9} & \perfcell{24.6} & \perfcell{30.9} & \perfcell{25.1} & \perfcell{52.2} & 34.0 & {11.0} & \perfcell{31.1} & \perfcell{25.1} & \perfcell{26.1} & \perfcell{24.6} & \perfcell{27.6} & \perfcell{41.7} & 29.4 & {6.5} \\
Baichuan Omni 1.5 & \perfcell{38.1} & \perfcell{28.0} & \perfcell{25.1} & \perfcell{31.7} & \perfcell{25.3} & \perfcell{61.2} & 34.9 & {13.8} & \perfcell{27.0} & \perfcell{25.2} & \perfcell{23.9} & \perfcell{26.9} & \perfcell{25.0} & \perfcell{52.2} & 30.0 & {10.9} \\
EchoInk-R1 & \perfcell{41.3} & \perfcell{27.2} & \perfcell{26.8} & \perfcell{34.0} & \perfcell{28.0} & \perfcell{62.2} & 36.6 & {13.7} & {\perfcell{38.2}} & \perfcell{26.2} & \perfcell{38.6} & \perfcell{31.1} & \perfcell{26.9} & \perfcell{61.6} & 37.1 & {13.1} \\
Qwen2.5-Omni & \perfcell{41.8} & \perfcell{31.2} & \perfcell{26.7} & \perfcell{34.4} & \perfcell{28.6} & \perfcell{67.8} & 38.4 & {15.3} & \perfcell{26.9} & \perfcell{28.7} & \perfcell{36.6} & \perfcell{25.6} & \perfcell{25.3} & \perfcell{50.8} & 32.3 & {10.0} \\  \midrule
Gemini 1.5 Pro & \perfcell{37.2} & \perfcell{31.2} & \perfcell{24.5} & \perfcell{51.4} & \perfcell{23.7} & \perfcell{72.8} & 40.1 & {19.0} & \perfcell{37.1} & \perfcell{27.2} & \perfcell{31.0} & \perfcell{47.3} & \perfcell{24.5} & \perfcell{55.7} & 37.1 & {12.2} \\
Gemini 2.0 Flash & \perfcell{45.2} & \textbf{\perfcell{43.1}} & \perfcell{29.2} & \perfcell{56.4} & \perfcell{33.5} & \perfcell{83.0} & 48.4 & {20.4} & \perfcell{51.8} & \perfcell{30.8} & \perfcell{38.6} & \perfcell{48.0} & \perfcell{27.4} & \perfcell{72.0} & 44.8 & {16.3} \\
Gemini 2.5 Flash & \textbf{\perfcell{45.6}} & \perfcell{31.4} & \perfcell{30.2} & \perfcell{71.2} & \perfcell{26.7} & \perfcell{83.2} & 48.0 & {23.8} & \perfcell{48.8} & \perfcell{39.6} & \perfcell{39.1} & \perfcell{51.4} & \perfcell{38.0} & \perfcell{74.6} & 48.6 & {13.9} \\
Gemini 2.5 Pro & \perfcell{41.0} & \perfcell{32.9} & \textbf{\perfcell{32.1}} & \textbf{\perfcell{75.8}} & \textbf{\perfcell{30.3}} & \textbf{\perfcell{88.3}} & \textbf{50.1} & {25.4} & \textbf{\perfcell{76.4}} & \textbf{\perfcell{54.4}} & \textbf{\perfcell{57.7}} & \textbf{\perfcell{55.4}} & \textbf{\perfcell{50.6}} & \textbf{\perfcell{70.6}} & \textbf{60.8} & {10.3} \\ \midrule
Human & 93.3 & 93.3 & 81.7 & 86.7 & 86.7 & 96.7 & 89.7 & 5.7 & 90.0 & 85.0 & 86.7 & 91.7 & 83.3 & 96.7 & 88.9 & 4.9 \\

\bottomrule
\end{tabular}
}

\resizebox{\textwidth}{!}{
\begin{tabular}{l|cccccc|c|c|cccccc|c|c}
\toprule
 & \multicolumn{8}{c|}{\textbf{Task 4 - Linguistic}} & \multicolumn{8}{c}{\textbf{Task 5 - External Knowledge}} \\
\cmidrule(lr){2-9} \cmidrule(lr){10-17}
\textbf{Model} & \textbf{A $\mapsto$ T} & \textbf{A $\mapsto$ V} & \textbf{T $\mapsto$ A} & \textbf{T $\mapsto$ V} & \textbf{V $\mapsto$ A} & \textbf{V $\mapsto$ T} & \textbf{\textit{Avg.}} & \textbf{\textit{Std.}} & \textbf{A $\mapsto$ T} & \textbf{A $\mapsto$ V} & \textbf{T $\mapsto$ A} & \textbf{T $\mapsto$ V} & \textbf{V $\mapsto$ A} & \textbf{V $\mapsto$ T} & \textbf{\textit{Avg.}} & \textbf{\textit{Std.}} \\
\midrule
PandaGPT & \perfcell{28.0} & \perfcell{24.3} & \perfcell{20.7} & \perfcell{24.7} & \perfcell{24.3} & \perfcell{31.3} & 25.5 & {3.6} & \perfcell{22.8} & \perfcell{23.9} & \perfcell{22.6} & \perfcell{23.6} & \perfcell{23.3} & \perfcell{22.6} & 23.1 & {0.5} \\
Unified-IO 2 & \perfcell{32.4} & \perfcell{27.5} & \perfcell{27.6} & \perfcell{27.9} & \perfcell{25.2} & \perfcell{41.7} & 30.4 & {6.0} & \perfcell{28.2} & \perfcell{24.2} & \perfcell{25.8} & \perfcell{27.1} & \perfcell{25.3} & \perfcell{29.9} & 26.8 & {2.1} \\
Unified-IO 2 XL & \perfcell{34.4} & \perfcell{31.7} & \perfcell{24.5} & \perfcell{28.8} & \perfcell{23.6} & \perfcell{41.8} & 30.8 & {6.8} & \perfcell{31.9} & \perfcell{27.9} & \perfcell{26.2} & \perfcell{28.6} & \perfcell{29.5} & \perfcell{32.9} & 29.5 & {2.5} \\
Unified-IO 2 XXL & \perfcell{39.9} & \perfcell{23.0} & \perfcell{25.5} & \perfcell{30.1} & \perfcell{22.3} & \perfcell{46.6} & 31.2 & {9.9} & \perfcell{38.4} & \perfcell{23.9} & \perfcell{35.3} & \perfcell{39.5} & \perfcell{29.7} & \perfcell{37.4} & 34.0 & {6.0} \\
VideoLLaMA 2 & \perfcell{50.3} & \perfcell{25.2} & \perfcell{24.2} & \perfcell{25.2} & \perfcell{24.1} & \perfcell{71.2} & 36.7 & {19.8} & \perfcell{42.9} & \perfcell{25.5} & \perfcell{23.9} & \perfcell{27.0} & \perfcell{24.4} & \perfcell{76.9} & 36.8 & {20.9} \\ 
% \midrule
VITA & \perfcell{52.2} & \perfcell{26.8} & \perfcell{47.1} & \perfcell{29.9} & \perfcell{47.9} & \perfcell{72.5} & 46.1 & {16.6} & \perfcell{38.5} & \perfcell{24.1} & \perfcell{24.2} & \perfcell{24.7} & \perfcell{22.6} & \perfcell{61.2} & 32.6 & {15.2} \\
Baichuan Omni 1.5 & \perfcell{77.0} & \perfcell{45.7} & \perfcell{65.8} & \perfcell{51.8} & \perfcell{58.7} & \perfcell{77.6} & 62.8 & {13.1} & \perfcell{54.3} & \perfcell{43.9} & \perfcell{41.9} & \perfcell{82.9} & \perfcell{33.7} & \perfcell{83.2} & 56.7 & {21.5} \\
EchoInk-R1 & \perfcell{86.0} & \perfcell{57.4} & \perfcell{74.6} & \perfcell{64.4} & \perfcell{70.1} & \perfcell{87.3} & 73.3 & {11.8} & \perfcell{83.3} & \perfcell{60.4} & \perfcell{72.7} & \perfcell{83.6} & \perfcell{56.6} & \perfcell{83.3} & 73.3 & {12.3} \\
Qwen2.5-Omni & \perfcell{85.6} & \perfcell{61.8} & \perfcell{73.6} & \perfcell{64.6} & \perfcell{71.5} & \perfcell{87.5} & 74.1 & {10.6} & \perfcell{83.0} & \perfcell{59.2} & \perfcell{70.7} & \perfcell{82.5} & \perfcell{58.6} & \perfcell{83.2} & 72.8 & {11.8} \\ \midrule
Gemini 1.5 Pro & \perfcell{86.2} & \perfcell{52.4} & \perfcell{72.3} & \perfcell{68.7} & \perfcell{70.7} & \perfcell{85.5} & 72.6 & {12.0} & \perfcell{62.3} & \perfcell{52.5} & \perfcell{70.2} & \perfcell{88.8} & \perfcell{52.3} & \perfcell{90.3} & 69.4 & {17.0} \\
Gemini 2.0 Flash & \perfcell{83.6} & \perfcell{57.5} & \perfcell{68.6} & \perfcell{67.3} & \perfcell{60.9} & \perfcell{83.4} & 70.2 & {11.1} & \perfcell{81.2} & \perfcell{68.3} & \perfcell{70.5} & \perfcell{93.1} & \perfcell{61.3} & \perfcell{94.2} & 78.1 & {13.6} \\
Gemini 2.5 Flash & \perfcell{84.1} & \textbf{\perfcell{68.3}} & \perfcell{70.9} & \perfcell{66.8} & \perfcell{64.4} & \perfcell{84.4} & 73.1 & {8.9} & \perfcell{82.0} & \perfcell{72.2} & \perfcell{81.7} & \perfcell{93.9} & \perfcell{74.5} & \perfcell{92.7} & 82.8 & {9.0} \\
Gemini 2.5 Pro & \textbf{\perfcell{84.9}} & \perfcell{67.5} & \textbf{\perfcell{75.5}} & \textbf{\perfcell{76.1}} & \textbf{\perfcell{65.8}} & \textbf{\perfcell{91.4}} & \textbf{76.8} & {9.9} & \textbf{\perfcell{90.3}} & \textbf{\perfcell{82.5}} & \textbf{\perfcell{88.2}} & \textbf{\perfcell{94.6}} & \textbf{\perfcell{84.8}} & \textbf{\perfcell{95.1}} & \textbf{89.3} & {5.1} \\ \midrule
Human & 89.2 & 96.7 & 97.5 & 93.3 & 91.7 & 95.0 & 93.9 & 2.8 & 96.7 & 88.3 & 98.3 & 98.3 & 85.0 & 96.7 & 93.9 & 5.8 \\
\bottomrule
\end{tabular}
}
\end{table}

\FloatBarrier

% \newpage
\section{Task specificed Model performance}\label{sec:taskperf}

% First part of the table
% \resizebox{0.9\textwidth}{!}{

% First part of the table
\subsection{Task 1: Perceptual Task}\label{sec:t1}

\scriptsize
\begin{longtable}{ll ccccc}
\caption{T1 (Perception) Results} \label{tab:t1_perception} \\
\toprule
\multicolumn{2}{c}{\textbf{Model}} & \multicolumn{5}{c}{\textbf{Perception Task}} \\
\cmidrule(lr){1-2} \cmidrule(lr){3-7}
\textbf{Model} & \textbf{Task} & \textbf{General} & \textbf{General - Hard} & \textbf{Scene} & \textbf{Instruments} & \textbf{Instruments-multi} \\
\midrule
\endfirsthead

\multicolumn{7}{c}%
{{\bfseries \tablename\ \thetable{} -- continued from previous page}} \\
\toprule
\multicolumn{2}{c}{\textbf{Model}} & \multicolumn{5}{c}{\textbf{Perception Task}} \\
\cmidrule(lr){1-2} \cmidrule(lr){3-7}
\textbf{Model} & \textbf{Task} & \textbf{General} & \textbf{General - Hard} & \textbf{Scene} & \textbf{Instruments} & \textbf{Instruments-multi} \\
\midrule
\endhead

\midrule \multicolumn{7}{r}{{Continued on next page}} \\
\endfoot

\bottomrule
\endlastfoot
\multirow{6}{*}{Gemini 2.5 Pro} 
& Audio $\mapsto$ Text & 81.05 & 71.39 & 67.20 & 47.75 & 44.09 \\
& Audio $\mapsto$ Vision & 76.26 & 65.25 & 64.60 & 44.30 & 36.60 \\
& Text $\mapsto$ Audio & 79.95 & 79.22 & 75.05 & 59.05 & 49.30 \\
& Text $\mapsto$ Vision & 98.90 & 97.87 & 90.80 & 97.90 & 99.80 \\
& Vision $\mapsto$ Audio & 88.73 & 79.35 & 84.40 & 61.92 & 48.79 \\
& Vision $\mapsto$ Text & 98.37 & 97.50 & 95.00 & 97.19 & 99.80 \\
\midrule
\multirow{6}{*}{Gemini 2.5 Flash} 
& Audio $\mapsto$ Text & 81.00 & 50.00 & 51.01 & 45.82 & 35.27 \\
& Audio $\mapsto$ Vision & 62.63 & 50.39 & 47.60 & 30.99 & 29.92 \\
& Text $\mapsto$ Audio & 79.80 & 59.13 & 57.34 & 37.90 & 32.99 \\
& Text $\mapsto$ Vision & 98.96 & 91.45 & 90.20 & 96.50 & 99.74 \\
& Vision $\mapsto$ Audio & 82.10 & 60.59 & 67.54 & 39.80 & 29.92 \\
& Vision $\mapsto$ Text & 98.39 & 96.62 & 92.60 & 89.88 & 97.27 \\
\midrule
\multirow{6}{*}{Gemini 2.0 Flash} 
& Audio $\mapsto$ Text & 81.10 & 62.07 & 54.00 & 47.05 & 39.80 \\
& Audio $\mapsto$ Vision & 67.45 & 51.68 & 49.80 & 31.50 & 24.80 \\
& Text $\mapsto$ Audio & 79.95 & 60.64 & 53.80 & 38.80 & 37.60 \\
& Text $\mapsto$ Vision & 98.95 & 91.45 & 80.40 & 96.90 & 95.60 \\
& Vision $\mapsto$ Audio & 82.50 & 66.45 & 53.00 & 37.90 & 35.40 \\
& Vision $\mapsto$ Text & 96.95 & 90.22 & 84.80 & 96.70 & 98.40 \\
\midrule
\multirow{6}{*}{Gemini 1.5 Pro} 
& Audio $\mapsto$ Text & 80.90 & 36.38 & 29.20 & 30.93 & 27.40 \\
& Audio $\mapsto$ Vision & 34.35 & 30.00 & 28.40 & 23.60 & 23.00 \\
& Text $\mapsto$ Audio & 80.25 & 45.88 & 41.80 & 31.10 & 26.20 \\
& Text $\mapsto$ Vision & 98.75 & 95.88 & 89.40 & 98.10 & 97.00 \\
& Vision $\mapsto$ Audio & 41.85 & 34.38 & 31.80 & 27.70 & 25.00 \\
& Vision $\mapsto$ Text & 95.10 & 94.62 & 87.80 & 98.90 & 100.00 \\
\midrule
\multirow{6}{*}{Qwen2.5 Omni} 
& Audio $\mapsto$ Text & 80.00 & 74.50 & 79.20 & 69.37 & 61.40 \\
& Audio $\mapsto$ Vision & 71.10 & 54.30 & 59.80 & 58.30 & 51.80 \\
& Text $\mapsto$ Audio & 81.20 & 69.90 & 78.80 & 67.40 & 48.80 \\
& Text $\mapsto$ Vision & 94.90 & 87.70 & 89.60 & 90.80 & 92.80 \\
& Vision $\mapsto$ Audio & 83.90 & 68.50 & 61.20 & 68.30 & 60.60 \\
& Vision $\mapsto$ Text & 97.50 & 87.00 & 88.00 & 91.80 & 95.80 \\
\midrule
\multirow{6}{*}{EchoInk} 
& Audio $\mapsto$ Text & 87.55 & 74.80 & 77.10 & 68.20 & 63.00 \\
& Audio $\mapsto$ Vision & 74.60 & 58.40 & 49.00 & 58.20 & 52.10 \\
& Text $\mapsto$ Audio & 84.57 & 66.40 & 79.40 & 69.14 & 46.80 \\
& Text $\mapsto$ Vision & 95.00 & 91.80 & 88.40 & 89.78 & 92.80 \\
& Vision $\mapsto$ Audio & 82.80 & 68.80 & 60.40 & 69.14 & 57.52 \\
& Vision $\mapsto$ Text & 96.00 & 95.20 & 87.80 & 92.38 & 95.79 \\
\midrule
\multirow{6}{*}{Baichuan Omni 1.5} 
& Audio $\mapsto$ Text & 55.85 & 44.05 & 46.40 & 36.44 & 31.00 \\
& Audio $\mapsto$ Vision & 44.45 & 37.43 & 43.20 & 29.60 & 26.60 \\
& Text $\mapsto$ Audio & 63.80 & 50.90 & 53.60 & 32.80 & 27.00 \\
& Text $\mapsto$ Vision & 97.35 & 88.10 & 81.20 & 84.50 & 88.00 \\
& Vision $\mapsto$ Audio & 68.25 & 53.75 & 58.80 & 38.40 & 32.20 \\
& Vision $\mapsto$ Text & 95.90 & 87.12 & 86.80 & 92.70 & 90.80 \\
\midrule
\multirow{6}{*}{VideoLLaMA 2} 
& Audio $\mapsto$ Text & 86.84 & 76.85 & 77.26 & 75.86 & 56.89 \\
& Audio $\mapsto$ Vision & 26.82 & 26.82 & 24.45 & 29.26 & 25.82 \\
& Text $\mapsto$ Audio & 30.69 & 28.21 & 28.89 & 28.45 & 25.07 \\
& Text $\mapsto$ Vision & 25.49 & 27.49 & 26.03 & 29.25 & 25.89 \\
& Vision $\mapsto$ Audio & 29.28 & 27.47 & 25.30 & 29.26 & 21.04 \\
& Vision $\mapsto$ Text & 97.05 & 91.48 & 87.45 & 89.40 & 92.23 \\
\midrule
\multirow{6}{*}{VITA} 
& Audio $\mapsto$ Text & 43.30 & 32.99 & 39.18 & 39.18 & 30.93 \\
& Audio $\mapsto$ Vision & 22.68 & 20.62 & 28.87 & 28.87 & 25.77 \\
& Text $\mapsto$ Audio & 28.96 & 24.24 & 24.92 & 28.28 & 28.62 \\
& Text $\mapsto$ Vision & 20.62 & 25.77 & 31.96 & 21.65 & 18.56 \\
& Vision $\mapsto$ Audio & 23.57 & 29.29 & 24.92 & 25.93 & 28.28 \\
& Vision $\mapsto$ Text & 64.95 & 73.20 & 58.76 & 74.23 & 74.23 \\
\midrule
\multirow{6}{*}{Unified IO 2} 
& Audio $\mapsto$ Text & 49.05 & 45.26 & 32.04 & 26.46 & 24.44 \\
& Audio $\mapsto$ Vision & 27.00 & 26.84 & 30.65 & 19.40 & 22.45 \\
& Text $\mapsto$ Audio & 26.68 & 25.86 & 25.27 & 27.64 & 26.20 \\
& Text $\mapsto$ Vision & 73.28 & 56.44 & 72.67 & 27.26 & 49.89 \\
& Vision $\mapsto$ Audio & 27.89 & 24.09 & 43.22 & 24.05 & 26.26 \\
& Vision $\mapsto$ Text & 55.83 & 44.21 & 32.81 & 48.66 & 41.86 \\
\midrule
\multirow{6}{*}{Unified IO 2 XL} 
& Audio $\mapsto$ Text & 76.64 & 71.68 & 57.82 & 33.68 & 26.87 \\
& Audio $\mapsto$ Vision & 28.04 & 25.21 & 34.89 & 22.29 & 29.22 \\
& Text $\mapsto$ Audio & 39.47 & 28.46 & 33.02 & 23.80 & 26.84 \\
& Text $\mapsto$ Vision & 81.89 & 68.28 & 69.87 & 22.09 & 53.29 \\
& Vision $\mapsto$ Audio & 26.46 & 24.43 & 35.05 & 24.80 & 27.03 \\
& Vision $\mapsto$ Text & 61.10 & 51.83 & 53.06 & 60.49 & 48.50 \\
\midrule
\multirow{6}{*}{Unified IO 2 XXL} 
& Audio $\mapsto$ Text & 83.63 & 71.20 & 45.88 & 41.45 & 32.83 \\
& Audio $\mapsto$ Vision & 29.07 & 23.28 & 27.41 & 27.09 & 27.87 \\
& Text $\mapsto$ Audio & 59.87 & 44.10 & 35.40 & 27.68 & 28.09 \\
& Text $\mapsto$ Vision & 86.07 & 73.08 & 71.29 & 36.07 & 54.44 \\
& Vision $\mapsto$ Audio & 28.66 & 29.81 & 24.82 & 24.01 & 26.07 \\
& Vision $\mapsto$ Text & 53.46 & 48.64 & 40.49 & 61.85 & 46.68 \\
\midrule
\multirow{2}{*}{PandaGPT}
& Audio $\mapsto$ Text & 25.03 & 28.80 & 24.49 & 24.30 & 19.99 \\
& Audio $\mapsto$ Vision & 26.52 & 27.37 & 24.63 & 24.89 & 20.15 \\
\multirow{4}{*}{PandaGPT}
& Text $\mapsto$ Audio & 25.40 & 29.65 & 24.25 & 24.77 & 20.08 \\
& Text $\mapsto$ Vision & 25.07 & 28.68 & 24.22 & 24.52 & 20.19 \\
& Vision $\mapsto$ Audio & 25.26 & 28.81 & 24.50 & 24.52 & 19.85 \\
& Vision $\mapsto$ Text & 25.26 & 28.70 & 24.64 & 24.90 & 20.05 \\
\end{longtable}

\subsection{Task 2: Spatial Reasoning}\label{sec:t2}

% T2 Spatial Task Table
\small
\begin{longtable}{ll ccc}
\caption{T2 (Spatial) Task Results} \label{tab:t2_spatial} \\
\toprule
\multicolumn{2}{c}{\textbf{Model}} & \multicolumn{3}{c}{\textbf{Spatial Task}} \\
\cmidrule(lr){1-2} \cmidrule(lr){3-5}
\textbf{Model} & \textbf{Task} & \textbf{Arrangement} & \textbf{Moving Direction} & \textbf{Indoor} \\
\midrule
\endfirsthead

\multicolumn{5}{c}%
{{\bfseries \tablename\ \thetable{} -- continued from previous page}} \\
\toprule
\multicolumn{2}{c}{\textbf{Model}} & \multicolumn{3}{c}{\textbf{Spatial Task}} \\
\cmidrule(lr){1-2} \cmidrule(lr){3-5}
\textbf{Model} & \textbf{Task} & \textbf{Arrangement} & \textbf{Moving Direction} & \textbf{Indoor} \\
\midrule
\endhead

\multicolumn{5}{r}{{Continued on next page}} \\
\endfoot

\bottomrule
\endlastfoot

\multirow{6}{*}{Gemini 2.5 Pro} 
& Audio $\mapsto$ Text & 28.82 & 69.39 & 24.87 \\
& Audio $\mapsto$ Vision & 24.73 & 40.65 & 33.38 \\
& Text $\mapsto$ Audio & 30.09 & 39.02 & 27.09 \\
& Text $\mapsto$ Vision & 95.70 & 58.85 & 72.73 \\
& Vision $\mapsto$ Audio & 29.01 & 38.10 & 23.86 \\
& Vision $\mapsto$ Text & 95.21 & 85.23 & 84.56 \\
\midrule
\multirow{6}{*}{Gemini 2.5 Flash} 
& Audio $\mapsto$ Text & 27.53 & 83.53 & 25.64 \\
& Audio $\mapsto$ Vision & 26.54 & 36.03 & 31.61 \\
& Text $\mapsto$ Audio & 25.81 & 35.34 & 29.37 \\
& Text $\mapsto$ Vision & 91.40 & 66.44 & 55.71 \\
& Vision $\mapsto$ Audio & 27.44 & 26.44 & 26.12 \\
& Vision $\mapsto$ Text & 91.40 & 84.05 & 74.25 \\
\midrule
\multirow{6}{*}{Gemini 2.0 Flash} 
& Audio $\mapsto$ Text & 28.82 & 82.71 & 24.10 \\
& Audio $\mapsto$ Vision & 26.45 & 37.58 & 35.38 \\
& Text $\mapsto$ Audio & 27.31 & 39.41 & 21.01 \\
& Text $\mapsto$ Vision & 67.53 & 66.99 & 34.62 \\
& Vision $\mapsto$ Audio & 25.81 & 45.78 & 28.86 \\
& Vision $\mapsto$ Text & 89.25 & 99.02 & 60.76 \\
\midrule
\multirow{6}{*}{Gemini 1.5 Pro} 
& Audio $\mapsto$ Text & 29.25 & 57.37 & 24.87 \\
& Audio $\mapsto$ Vision & 27.10 & 32.87 & 33.59 \\
& Text $\mapsto$ Audio & 19.25 & 34.26 & 20.00 \\
& Text $\mapsto$ Vision & 64.30 & 50.80 & 39.23 \\
& Vision $\mapsto$ Audio & 23.66 & 21.82 & 25.57 \\
& Vision $\mapsto$ Text & 95.48 & 80.00 & 43.04 \\
\midrule
\multirow{6}{*}{Qwen2.5 Omni} 
& Audio $\mapsto$ Text & 21.29 & 75.28 & 28.89 \\
& Audio $\mapsto$ Vision & 28.60 & 35.83 & 29.23 \\
& Text $\mapsto$ Audio & 20.22 & 31.52 & 28.35 \\
& Text $\mapsto$ Vision & 45.38 & 26.98 & 30.77 \\
& Vision $\mapsto$ Audio & 23.87 & 34.69 & 27.34 \\
& Vision $\mapsto$ Text & 80.86 & 81.63 & 41.01 \vspace{1em}\\ 
\midrule
\multirow{6}{*}{EchoInk} 
& Audio $\mapsto$ Text & 27.79 & 61.62 & 34.34 \\
& Audio $\mapsto$ Vision & 24.97 & 25.59 & 30.98 \\
& Text $\mapsto$ Audio & 26.60 & 28.96 & 24.92 \\
& Text $\mapsto$ Vision & 46.80 & 25.59 & 29.63 \\
& Vision $\mapsto$ Audio & 24.88 & 31.31 & 27.95 \\
& Vision $\mapsto$ Text & 80.13 & 61.62 & 44.78 \\
\midrule
\multirow{6}{*}{Baichuan Omni 1.5} 
& Audio $\mapsto$ Text & 28.39 & 71.43 & 14.36 \\
& Audio $\mapsto$ Vision & 28.17 & 28.51 & 27.18 \\
& Text $\mapsto$ Audio & 22.37 & 27.21 & 25.57 \\
& Text $\mapsto$ Vision & 35.70 & 36.32 & 23.08 \\
& Vision $\mapsto$ Audio & 25.38 & 22.95 & 27.59 \\
& Vision $\mapsto$ Text & 71.40 & 82.95 & 29.37 \vspace{1em} \\
\midrule
\multirow{6}{*}{VideoLLaMA 2} 
& Audio $\mapsto$ Text & 31.40 & 62.44 & 37.76 \\
% \midrule
% \multirow{5}{*}{VideoLLaMA 2} 
& Audio $\mapsto$ Vision & 27.40 & 27.75 & 28.22 \\
& Text $\mapsto$ Audio & 26.76 & 27.04 & 19.53 \\
& Text $\mapsto$ Vision & 27.36 & 27.01 & 28.25 \\
& Vision $\mapsto$ Audio & 25.63 & 29.28 & 20.77 \\
& Vision $\mapsto$ Text & 46.96 & 84.21 & 31.70 \\
\midrule
\multirow{6}{*}{VITA} 
& Audio $\mapsto$ Text & 29.90 & 77.32 & 19.59 \\
& Audio $\mapsto$ Vision & 30.93 & 26.80 & 28.87 \\
& Text $\mapsto$ Audio & 23.23 & 25.59 & 25.00 \\
& Text $\mapsto$ Vision & 29.90 & 31.96 & 30.93 \\
& Vision $\mapsto$ Audio & 24.92 & 25.59 & 24.66 \\
& Vision $\mapsto$ Text & 57.73 & 55.67 & 43.30 \\
\midrule
\multirow{6}{*}{Unified IO 2} 
& Audio $\mapsto$ Text & 23.03 & 20.47 & 34.40 \\
& Audio $\mapsto$ Vision & 21.98 & 17.20 & 22.89 \\
& Text $\mapsto$ Audio & 23.50 & 20.69 & 22.87 \\
& Text $\mapsto$ Vision & 25.63 & 24.03 & 25.22 \\
& Vision $\mapsto$ Audio & 24.09 & 17.32 & 27.86 \\
& Vision $\mapsto$ Text & 28.60 & 24.10 & 21.49 \\
\midrule
\multirow{6}{*}{Unified IO 2 XL} 
& Audio $\mapsto$ Text & 23.09 & 28.42 & 22.88 \\
& Audio $\mapsto$ Vision & 22.20 & 20.09 & 26.75 \\
& Text $\mapsto$ Audio & 24.82 & 22.92 & 29.70 \\
& Text $\mapsto$ Vision & 24.18 & 24.25 & 29.56 \\
& Vision $\mapsto$ Audio & 24.12 & 21.78 & 32.17 \\
& Vision $\mapsto$ Text & 27.41 & 24.10 & 21.93 \\
\midrule
\multirow{6}{*}{Unified IO 2 XXL} 
& Audio $\mapsto$ Text & 22.58 & 30.07 & 36.18 \\
& Audio $\mapsto$ Vision & 24.54 & 24.02 & 22.37 \\
& Text $\mapsto$ Audio & 25.85 & 38.11 & 28.71 \\
& Text $\mapsto$ Vision & 25.39 & 30.02 & 21.10 \\
& Vision $\mapsto$ Audio & 25.45 & 28.33 & 34.77 \\
& Vision $\mapsto$ Text & 30.36 & 30.91 & 30.80 \\
\midrule
\multirow{6}{*}{PandaGPT} 
& Audio $\mapsto$ Text & 25.42 & 25.62 & 25.44 \\
& Audio $\mapsto$ Vision & 27.22 & 25.63 & 26.91 \\
& Text $\mapsto$ Audio & 27.06 & 25.58 & 25.27 \\
& Text $\mapsto$ Vision & 27.01 & 25.95 & 28.57 \\
& Vision $\mapsto$ Audio & 27.16 & 25.53 & 24.57 \\
& Vision $\mapsto$ Text & 21.19 & 25.72 & 22.34 \\
\end{longtable}

\subsection{Task 3: Temporal Reasoning}\label{sec:t3}

% T3 Temporal Task Table
\small
\begin{longtable}{ll ccc}
\caption{T3 (Temporal) Task Results} \label{tab:t3_temporal} \\
\toprule
\multicolumn{2}{c}{\textbf{Model}} & \multicolumn{3}{c}{\textbf{Temporal Task}} \\
\cmidrule(lr){1-2} \cmidrule(lr){3-5}
\textbf{Model} & \textbf{Task} & \textbf{Order} & \textbf{Counting} & \textbf{Calculation} \\
\midrule
\endfirsthead

\multicolumn{5}{c}%
{{\bfseries \tablename\ \thetable{} -- continued from previous page}} \\
\toprule
\multicolumn{2}{c}{\textbf{Model}} & \multicolumn{3}{c}{\textbf{Temporal Task}} \\
\cmidrule(lr){1-2} \cmidrule(lr){3-5}
\textbf{Model} & \textbf{Task} & \textbf{Order} & \textbf{Counting} & \textbf{Calculation} \\
\midrule
\endhead

\multicolumn{5}{r}{{Continued on next page}} \\
\endfoot

\bottomrule
\endlastfoot

\multirow{6}{*}{Gemini 2.5 Pro} 
& Audio $\mapsto$ Text & 96.18 & 57.36 & 75.78 \\
& Audio $\mapsto$ Vision & 95.38 & 37.88 & 29.87 \\
& Text $\mapsto$ Audio & 95.39 & 50.00 & 27.60 \\
& Text $\mapsto$ Vision & 99.80 & 35.85 & 30.63 \\
& Vision $\mapsto$ Audio & 96.35 & 34.70 & 20.65 \\
& Vision $\mapsto$ Text & 99.80 & 40.58 & 71.46 \\
\midrule
\multirow{6}{*}{Gemini 2.5 Flash} 
& Audio $\mapsto$ Text & 41.40 & 49.60 & 55.40 \\
& Audio $\mapsto$ Vision & 58.99 & 33.07 & 26.88 \\
& Text $\mapsto$ Audio & 61.00 & 29.40 & 27.00 \\
& Text $\mapsto$ Vision & 99.15 & 29.45 & 25.51 \\
& Vision $\mapsto$ Audio & 63.39 & 26.58 & 24.13 \\
& Vision $\mapsto$ Text & 99.20 & 53.37 & 71.22 \\
\midrule
\multirow{6}{*}{Gemini 2.0 Flash} 
& Audio $\mapsto$ Text & 43.60 & 52.60 & 59.20 \\
& Audio $\mapsto$ Vision & 33.40 & 30.17 & 28.93 \\
& Text $\mapsto$ Audio & 61.40 & 28.80 & 25.60 \\
& Text $\mapsto$ Vision & 81.40 & 33.33 & 29.16 \\
& Vision $\mapsto$ Audio & 33.40 & 28.22 & 20.65 \\
& Vision $\mapsto$ Text & 99.20 & 57.87 & 58.90 \\
\midrule
\multirow{6}{*}{Gemini 1.5 Pro} 
& Audio $\mapsto$ Text & 34.40 & 30.00 & 47.00 \\
& Audio $\mapsto$ Vision & 32.00 & 24.44 & 25.10 \\
& Text $\mapsto$ Audio & 38.60 & 30.20 & 24.20 \\
& Text $\mapsto$ Vision & 82.00 & 33.88 & 26.14 \\
& Vision $\mapsto$ Audio & 27.60 & 23.87 & 21.88 \\
& Vision $\mapsto$ Text & 98.40 & 25.26 & 43.56 \\
\midrule
\multirow{6}{*}{Qwen2.5 Omni} 
& Audio $\mapsto$ Text & 28.20 & 25.80 & 26.60 \\
& Audio $\mapsto$ Vision & 34.80 & 22.22 & 28.96 \\
& Text $\mapsto$ Audio & 63.80 & 19.40 & 26.60 \\
& Text $\mapsto$ Vision & 24.80 & 22.90 & 28.96 \\
& Vision $\mapsto$ Audio & 26.40 & 23.57 & 25.93 \\
& Vision $\mapsto$ Text & 85.00 & 41.41 & 25.93 \\
\midrule
\multirow{6}{*}{EchoInk} 
& Audio $\mapsto$ Text & 35.00 & 48.48 & 30.98 \\
& Audio $\mapsto$ Vision & 30.98 & 23.57 & 23.91 \\
& Text $\mapsto$ Audio & 68.69 & 21.89 & 25.25 \\
& Text $\mapsto$ Vision & 43.10 & 22.56 & 27.61 \\
& Vision $\mapsto$ Audio & 31.99 & 23.57 & 25.25 \\
& Vision $\mapsto$ Text & 93.60 & 46.80 & 44.44 \\
\midrule
\multirow{6}{*}{Baichuan Omni 1.5} 
& Audio $\mapsto$ Text & 23.00 & 34.40 & 23.60 \\
& Audio $\mapsto$ Vision & 23.80 & 25.74 & 25.99 \\
& Text $\mapsto$ Audio & 23.40 & 23.00 & 25.40 \\
& Text $\mapsto$ Vision & 25.80 & 26.23 & 28.77 \\
& Vision $\mapsto$ Audio & 25.20 & 28.34 & 21.47 \\
& Vision $\mapsto$ Text & 70.20 & 53.18 & 33.13 \\
\midrule
\multirow{6}{*}{VideoLLaMA 2} 
& Audio $\mapsto$ Text & 25.82 & 35.90 & 31.23 \\
& Audio $\mapsto$ Vision & 25.23 & 25.80 & 24.03 \\
& Text $\mapsto$ Audio & 34.29 & 22.09 & 26.70 \\
& Text $\mapsto$ Vision & 26.66 & 26.06 & 24.90 \\
& Vision $\mapsto$ Audio & 27.03 & 23.64 & 26.67 \\
& Vision $\mapsto$ Text & 50.40 & 32.44 & 36.67 \\
\midrule
\multirow{6}{*}{VITA} 
& Audio $\mapsto$ Text & 26.26 & 38.14 & 28.87 \\
& Audio $\mapsto$ Vision & 16.49 & 31.17 & 27.52 \\
& Text $\mapsto$ Audio & 26.80 & 27.61 & 23.91 \\
& Text $\mapsto$ Vision & 22.68 & 25.62 & 25.58 \\
& Vision $\mapsto$ Audio & 28.62 & 26.71 & 27.59 \\
& Vision $\mapsto$ Text & 42.27 & 49.66 & 33.10 \\
\midrule
\multirow{6}{*}{Unified IO 2} 
& Audio $\mapsto$ Text & 24.28 & 18.25 & 25.44 \\
& Audio $\mapsto$ Vision & 21.50 & 22.61 & 23.03 \\
& Text $\mapsto$ Audio & 30.02 & 23.46 & 21.89 \\
& Text $\mapsto$ Vision & 25.25 & 24.85 & 22.86 \\
& Vision $\mapsto$ Audio & 25.46 & 26.29 & 25.65 \\
& Vision $\mapsto$ Text & 27.68 & 16.25 & 23.37 \\
\midrule
\multirow{6}{*}{Unified IO 2 XL} 
& Audio $\mapsto$ Text & 24.65 & 24.63 & 17.47 \\
& Audio $\mapsto$ Vision & 26.03 & 30.21 & 17.39 \\
& Text $\mapsto$ Audio & 27.52 & 28.83 & 30.02 \\
& Text $\mapsto$ Vision & 25.09 & 19.16 & 22.17 \\
& Vision $\mapsto$ Audio & 22.64 & 24.92 & 30.57 \\
& Vision $\mapsto$ Text & 37.30 & 36.42 & 24.44 \\
\midrule
\multirow{6}{*}{Unified IO 2 XXL} 
& Audio $\mapsto$ Text & 24.41 & 26.81 & 21.62 \\
& Audio $\mapsto$ Vision & 25.26 & 29.68 & 27.17 \\
& Text $\mapsto$ Audio & 28.83 & 22.43 & 24.61 \\
& Text $\mapsto$ Vision & 23.70 & 37.78 & 27.37 \\
& Vision $\mapsto$ Audio & 23.63 & 24.69 & 27.28 \\
& Vision $\mapsto$ Text & 41.69 & 38.50 & 22.95 \\
\midrule
\multirow{6}{*}{Panda} 
& Audio $\mapsto$ Text & 25.85 & 16.77 & 23.17 \\
& Audio $\mapsto$ Vision & 26.06 & 22.60 & 27.31 \\
& Text $\mapsto$ Audio & 25.72 & 22.81 & 25.80 \\
& Text $\mapsto$ Vision & 26.31 & 22.77 & 29.02 \\
& Vision $\mapsto$ Audio & 26.10 & 22.77 & 24.59 \\
& Vision $\mapsto$ Text & 25.51 & 22.94 & 23.37 \\
\end{longtable}

\subsection{Task 4: Linguistic Task}\label{sec:t4}

% T4 Speech Task Table
\small
\begin{longtable}{ll ccc}
\caption{T4 Linguistic Task Results} \label{tab:t4_speech} \\
\toprule
\multicolumn{2}{c}{\textbf{Model}} & \multicolumn{3}{c}{\textbf{Linguistic Task}} \\
\cmidrule(lr){1-2} \cmidrule(lr){3-5}
\textbf{Model} & \textbf{Task} & \textbf{Recognition} & \textbf{Translation} & \textbf{Emotion} \\
\midrule
\endfirsthead

\multicolumn{5}{c}%
{{\bfseries \tablename\ \thetable{} -- continued from previous page}} \\
\toprule
\multicolumn{2}{c}{\textbf{Model}} & \multicolumn{3}{c}{\textbf{Linguistic Task}} \\
\cmidrule(lr){1-2} \cmidrule(lr){3-5}
\textbf{Model} & \textbf{Task} & \textbf{Recognition} & \textbf{Translation} & \textbf{Emotion} \\
\midrule
\endhead

\midrule \multicolumn{5}{r}{{Continued on next page}} \\
\endfoot

\bottomrule
\endlastfoot

\multirow{6}{*}{Gemini 2.5 Pro} 
& Audio $\mapsto$ Text & 97.16 & 96.58 & 60.86 \\
& Audio $\mapsto$ Vision & 91.65 & 67.95 & 42.75 \\
& Text $\mapsto$ Audio & 80.35 & 81.62 & 64.51 \\
& Text $\mapsto$ Vision & 93.58 & 67.38 & 67.31 \\
& Vision $\mapsto$ Audio & 80.81 & 73.22 & 43.43 \\
& Vision $\mapsto$ Text & 99.54 & 100.00 & 74.54 \\
\midrule
\multirow{6}{*}{Gemini 2.5 Flash} 
& Audio $\mapsto$ Text & 94.05 & 97.44 & 60.86 \\
& Audio $\mapsto$ Vision & 68.01 & 93.30 & 43.67 \\
& Text $\mapsto$ Audio & 76.92 & 81.34 & 54.43 \\
& Text $\mapsto$ Vision & 72.88 & 67.24 & 60.14 \\
& Vision $\mapsto$ Audio & 74.95 & 72.93 & 45.22 \\
& Vision $\mapsto$ Text & 99.40 & 96.72 & 57.14 \\
\midrule
\multirow{6}{*}{Gemini 2.0 Flash} 
& Audio $\mapsto$ Text & 92.86 & 97.29 & 60.57 \\
& Audio $\mapsto$ Vision & 68.30 & 67.66 & 36.43 \\
& Text $\mapsto$ Audio & 69.79 & 81.20 & 54.86 \\
& Text $\mapsto$ Vision & 73.92 & 67.66 & 60.43 \\
& Vision $\mapsto$ Audio & 66.52 & 73.08 & 43.00 \\
& Vision $\mapsto$ Text & 96.43 & 97.15 & 56.71 \\
\midrule
\multirow{6}{*}{Gemini 1.5 Pro} 
& Audio $\mapsto$ Text & 94.94 & 97.15 & 60.43 \\
& Audio $\mapsto$ Vision & 73.96 & 46.72 & 36.57 \\
& Text $\mapsto$ Audio & 83.33 & 80.91 & 52.57 \\
& Text $\mapsto$ Vision & 76.93 & 66.81 & 62.43 \\
& Vision $\mapsto$ Audio & 80.80 & 92.02 & 39.20 \\
& Vision $\mapsto$ Text & 96.73 & 96.44 & 63.29 \\
\midrule
\multirow{6}{*}{Qwen2.5 Omni} 
& Audio $\mapsto$ Text & 94.64 & 96.72 & 65.29 \\
& Audio $\mapsto$ Vision & 62.95 & 73.36 & 48.94 \\
& Text $\mapsto$ Audio & 81.25 & 86.75 & 52.71 \\
& Text $\mapsto$ Vision & 65.03 & 69.09 & 59.79 \\
& Vision $\mapsto$ Audio & 82.44 & 88.60 & 43.57 \\
& Vision $\mapsto$ Text & 97.17 & 97.72 & 67.57 \\
\midrule
\multirow{3}{*}{EchoInk} 
& Audio $\mapsto$ Text & 92.93 & 95.96 & 69.02 \\
& Audio $\mapsto$ Vision & 64.98 & 71.38 & 35.69 \\
& Text $\mapsto$ Audio & 80.47 & 81.48 & 61.95 \\
& Text $\mapsto$ Vision & 68.35 & 67.68 & 57.24 \\
& Vision $\mapsto$ Audio & 81.48 & 85.86 & 43.10 \\
& Vision $\mapsto$ Text & 96.63 & 97.31 & 68.01 \\
\midrule
\multirow{6}{*}{Baichuan Omni 1.5} 
& Audio $\mapsto$ Text & 87.05 & 96.01 & 48.00 \\
& Audio $\mapsto$ Vision & 55.36 & 56.55 & 25.25 \\
& Text $\mapsto$ Audio & 64.29 & 84.94 & 48.29 \\
& Text $\mapsto$ Vision & 55.95 & 52.56 & 46.99 \\
& Vision $\mapsto$ Audio & 65.03 & 84.06 & 27.14 \\
& Vision $\mapsto$ Text & 92.56 & 96.72 & 43.43 \\
\midrule
\multirow{6}{*}{VideoLLaMA 2} 
& Audio $\mapsto$ Text & 69.04 & 67.40 & 14.48 \\
& Audio $\mapsto$ Vision & 24.82 & 26.00 & 24.68 \\
& Text $\mapsto$ Audio & 22.82 & 22.02 & 27.68 \\
& Text $\mapsto$ Vision & 25.03 & 25.80 & 24.65 \\
& Vision $\mapsto$ Audio & 24.07 & 23.25 & 25.01 \\
& Vision $\mapsto$ Text & 83.86 & 86.80 & 43.00 \\
\midrule
\multirow{6}{*}{VITA} 
& Audio $\mapsto$ Text & 39.18 & 73.20 & 44.33 \\
& Audio $\mapsto$ Vision & 24.74 & 24.74 & 30.93 \\
& Text $\mapsto$ Audio & 39.73 & 55.56 & 46.13 \\
& Text $\mapsto$ Vision & 30.93 & 25.77 & 32.99 \\
& Vision $\mapsto$ Audio & 53.87 & 61.95 & 27.95 \\
& Vision $\mapsto$ Text & 86.60 & 88.66 & 42.27 \\
\midrule
\multirow{6}{*}{Unified IO 2} 
& Audio $\mapsto$ Text & 62.01 & 14.06 & 21.05 \\
& Audio $\mapsto$ Vision & 35.66 & 20.90 & 25.83 \\
& Text $\mapsto$ Audio & 26.60 & 26.36 & 29.85 \\
& Text $\mapsto$ Vision & 25.89 & 26.00 & 31.82 \\
& Vision $\mapsto$ Audio & 24.24 & 25.14 & 26.27 \\
& Vision $\mapsto$ Text & 66.06 & 18.90 & 40.01 \\
\midrule
\multirow{6}{*}{Unified IO 2 XL} 
& Audio $\mapsto$ Text & 69.63 & 17.26 & 16.29 \\
& Audio $\mapsto$ Vision & 45.46 & 26.28 & 23.28 \\
& Text $\mapsto$ Audio & 27.82 & 23.75 & 21.90 \\
& Text $\mapsto$ Vision & 30.65 & 25.26 & 30.47 \\
& Vision $\mapsto$ Audio & 25.07 & 23.70 & 21.88 \\
& Vision $\mapsto$ Text & 75.27 & 23.23 & 27.02 \\
\midrule
\multirow{6}{*}{Unified IO 2 XXL} 
& Audio $\mapsto$ Text & 72.67 & 17.63 & 29.46 \\
& Audio $\mapsto$ Vision & 18.23 & 27.43 & 23.24 \\
& Text $\mapsto$ Audio & 23.04 & 25.97 & 27.47 \\
& Text $\mapsto$ Vision & 31.09 & 27.84 & 31.42 \\
& Vision $\mapsto$ Audio & 19.43 & 23.31 & 24.06 \\
& Vision $\mapsto$ Text & 78.04 & 26.88 & 34.88 \\
\midrule
\multirow{6}{*}{PandaGPT} 
& Audio $\mapsto$ Text & 27.12 & 28.83 & 28.03 \\
& Audio $\mapsto$ Vision & 22.38 & 22.23 & 28.23 \\
& Text $\mapsto$ Audio & 22.06 & 18.88 & 21.20 \\
& Text $\mapsto$ Vision & 22.10 & 24.96 & 27.04 \\
& Vision $\mapsto$ Audio & 22.55 & 22.40 & 27.99 \\
& Vision $\mapsto$ Text & 33.96 & 32.67 & 27.20 \\
\end{longtable}

\subsection{Task 5: External Knowledge}\label{sec:t5}

% T5 External Task Table
\small
\begin{longtable}{ll ccc}
\caption{T5 (External) Task Results} \label{tab:t5_external} \\
\toprule
\multicolumn{2}{c}{\textbf{Model}} & \multicolumn{3}{c}{\textbf{External Task}} \\
\cmidrule(lr){1-2} \cmidrule(lr){3-5}
\textbf{Model} & \textbf{Task} & \textbf{Genre} & \textbf{Movie} & \textbf{Singer} \\
\midrule
\endfirsthead

\multicolumn{5}{c}%
{{\bfseries \tablename\ \thetable{} -- continued from previous page}} \\
\toprule
\multicolumn{2}{c}{\textbf{Model}} & \multicolumn{3}{c}{\textbf{External Task}} \\
\cmidrule(lr){1-2} \cmidrule(lr){3-5}
\textbf{Model} & \textbf{Task} & \textbf{Genre} & \textbf{Movie} & \textbf{Singer} \\
\midrule
\endhead

\midrule \multicolumn{5}{r}{{Continued on next page}} \\
\endfoot

\bottomrule
\endlastfoot

\multirow{6}{*}{Gemini 2.5 Pro} 
& Audio $\mapsto$ Text & 83.28 & 93.00 & 94.67 \\
& Audio $\mapsto$ Vision & 74.80 & 89.90 & 82.67 \\
& Text $\mapsto$ Audio & 78.16 & 94.50 & 91.95 \\
& Text $\mapsto$ Vision & 85.76 & 97.99 & 100.00 \\
& Vision $\mapsto$ Audio & 72.42 & 92.00 & 90.00 \\
& Vision $\mapsto$ Text & 88.95 & 96.45 & 100.00 \\
\midrule
\multirow{6}{*}{Gemini 2.5 Flash} 
& Audio $\mapsto$ Text & 83.78 & 93.00 & 69.13 \\
& Audio $\mapsto$ Vision & 63.36 & 82.41 & 70.92 \\
& Text $\mapsto$ Audio & 78.56 & 90.45 & 76.00 \\
& Text $\mapsto$ Vision & 85.00 & 97.99 & 98.67 \\
& Vision $\mapsto$ Audio & 63.96 & 88.32 & 71.33 \\
& Vision $\mapsto$ Text & 86.34 & 98.00 & 93.71 \\
\midrule
\multirow{6}{*}{Gemini 2.0 Flash} 
& Audio $\mapsto$ Text & 83.50 & 88.00 & 72.00 \\
& Audio $\mapsto$ Vision & 62.40 & 86.50 & 56.00 \\
& Text $\mapsto$ Audio & 78.46 & 82.50 & 50.67 \\
& Text $\mapsto$ Vision & 84.50 & 98.00 & 96.67 \\
& Vision $\mapsto$ Audio & 66.43 & 79.50 & 38.00 \\
& Vision $\mapsto$ Text & 87.50 & 95.00 & 100.00 \\
\midrule
\multirow{6}{*}{Gemini 1.5 Pro} 
& Audio $\mapsto$ Text & 61.70 & 78.00 & 47.33 \\
& Audio $\mapsto$ Vision & 42.90 & 74.50 & 40.00 \\
& Text $\mapsto$ Audio & 63.53 & 84.50 & 62.67 \\
& Text $\mapsto$ Vision & 82.10 & 95.50 & 88.67 \\
& Vision $\mapsto$ Audio & 45.59 & 74.00 & 37.33 \\
& Vision $\mapsto$ Text & 87.10 & 95.00 & 88.67 \\
\midrule
\multirow{6}{*}{Qwen2.5 Omni} 
& Audio $\mapsto$ Text & 89.50 & 79.50 & 80.00 \\
& Audio $\mapsto$ Vision & 61.40 & 67.50 & 48.67 \\
& Text $\mapsto$ Audio & 85.65 & 70.50 & 56.00 \\
& Text $\mapsto$ Vision & 74.20 & 94.50 & 78.67 \\
& Vision $\mapsto$ Audio & 81.82 & 60.50 & 33.33 \\
& Vision $\mapsto$ Text & 79.00 & 92.50 & 78.00 \\
\midrule
\multirow{6}{*}{EchoInk} 
& Audio $\mapsto$ Text & 87.54 & 82.50 & 80.00 \\
& Audio $\mapsto$ Vision & 61.95 & 68.00 & 51.33 \\
& Text $\mapsto$ Audio & 84.51 & 73.00 & 60.67 \\
& Text $\mapsto$ Vision & 77.78 & 93.00 & 80.00 \\
& Vision $\mapsto$ Audio & 62.63 & 64.50 & 42.67 \\
& Vision $\mapsto$ Text & 79.12 & 93.50 & 77.33 \\
\midrule
\multirow{6}{*}{Baichuan Omni 1.5} 
& Audio $\mapsto$ Text & 65.60 & 56.00 & 41.33 \\
& Audio $\mapsto$ Vision & 45.30 & 54.50 & 32.00 \\
& Text $\mapsto$ Audio & 25.75 & 60.00 & 40.00 \\
& Text $\mapsto$ Vision & 77.00 & 94.50 & 77.33 \\
& Vision $\mapsto$ Audio & 27.15 & 46.50 & 27.33 \\
& Vision $\mapsto$ Text & 81.20 & 94.50 & 74.00 \\
\midrule
\multirow{4}{*}{VideoLLaMA 2} 
& Audio $\mapsto$ Text & 62.60 & 26.59 & 39.38 \\
& Audio $\mapsto$ Vision & 26.23 & 23.56 & 26.72 \\
& Text $\mapsto$ Audio & 24.85 & 21.59 & 25.34 \\
& Text $\mapsto$ Vision & 26.40 & 28.55 & 26.09 \\
& Vision $\mapsto$ Audio & 25.67 & 23.56 & 24.10 \\
& Vision $\mapsto$ Text & 68.27 & 80.55 & 82.02 \\
\midrule
\multirow{6}{*}{VITA} 
& Audio $\mapsto$ Text & 46.39 & 40.21 & 28.87 \\
& Audio $\mapsto$ Vision & 20.62 & 26.80 & 24.74 \\
& Text $\mapsto$ Audio & 21.89 & 25.50 & 25.33 \\
& Text $\mapsto$ Vision & 20.62 & 31.96 & 21.65 \\
& Vision $\mapsto$ Audio & 23.23 & 22.00 & 22.67 \\
& Vision $\mapsto$ Text & 47.42 & 81.44 & 54.64 \\
\midrule
\multirow{6}{*}{Unified IO 2} 
& Audio $\mapsto$ Text & 31.83 & 22.53 & 30.09 \\
& Audio $\mapsto$ Vision & 22.30 & 29.03 & 21.40 \\
    & Text $\mapsto$ Audio & 26.25 & 24.51 & 26.71 \\
& Text $\mapsto$ Vision & 34.46 & 26.03 & 20.71 \\
& Vision $\mapsto$ Audio & 25.45 & 20.57 & 30.01 \\
& Vision $\mapsto$ Text & 27.90 & 34.59 & 27.33 \\
\midrule
\multirow{6}{*}{Unified IO 2 XL} 
& Audio $\mapsto$ Text & 36.80 & 27.52 & 31.40 \\
& Audio $\mapsto$ Vision & 29.23 & 29.09 & 25.40 \\
& Text $\mapsto$ Audio & 24.12 & 25.09 & 29.34 \\
& Text $\mapsto$ Vision & 34.41 & 24.57 & 26.68 \\
& Vision $\mapsto$ Audio & 26.51 & 32.55 & 29.41 \\
& Vision $\mapsto$ Text & 24.86 & 35.76 & 38.05 \\
\midrule
\multirow{6}{*}{Unified IO 2 XXL} 
& Audio $\mapsto$ Text & 57.68 & 22.71 & 34.70 \\
& Audio $\mapsto$ Vision & 26.83 & 20.56 & 24.42 \\
& Text $\mapsto$ Audio & 47.92 & 26.52 & 31.43 \\
& Text $\mapsto$ Vision & 51.85 & 24.01 & 42.75 \\
& Vision $\mapsto$ Audio & 25.06 & 30.57 & 33.40 \\
& Vision $\mapsto$ Text & 28.20 & 36.55 & 47.36 \\
\midrule
\multirow{6}{*}{Panda} 
& Audio $\mapsto$ Text & 25.77 & 21.32 & 21.24 \\
& Audio $\mapsto$ Vision & 25.74 & 24.49 & 21.42 \\
& Text $\mapsto$ Audio & 22.11 & 25.18 & 20.37 \\
& Text $\mapsto$ Vision & 24.63 & 24.64 & 21.40 \\
& Vision $\mapsto$ Audio & 23.93 & 24.58 & 21.29 \\
& Vision $\mapsto$ Text & 26.32 & 21.07 & 20.39 \\
\end{longtable}

\section{Evaluation Cost}\label{sec:cost}
\revise{
We provide a detailed evaluation cost section as a reference of usage. We evaluate on the full version (60k sample) of XModBench, API-based models we test \textbf{Gemini 2.5 Pro}, we report the \emph{token usage} for evaluating the overall benchmark and each task family . For open-source models we report \textbf{Qwen2.5-Omni}, we report the \emph{evaluation runing time}, using with eight \texttt{A6000} GPUs and each GPU run one process.}

\begin{table}[h!]
\centering
\small
\caption{Evaluation cost estimation for models across the five task families and the full benchmark.}
\label{tab:evaluation_cost}
\setlength{\tabcolsep}{8pt}
\begin{tabular}{l|ccccc|c}
\toprule
\textbf{Model} & \textbf{Perc.} & \textbf{Spat.} & \textbf{Temp.} & \textbf{Ling.} & \textbf{Knwl.} & \textbf{Total} \\
\midrule
% \multicolumn{7}{l}{\textbf{Token usage}} \\
Gemini 2.5 Pro (\textit{Token usage}) 
  & 26.0M & 13.5M & 25.1M & 4.3M & 14.0M & 82.9M \\
\midrule
% \multicolumn{7}{l}{\textbf{Evaluation time (hours)}} \\
Qwen2.5-Omni (\textit{Hours})
  & 6.3 & 1.4 & 1.4 & 1.4 & 2.1 & 12.7 \\
\bottomrule
\end{tabular}
\end{table}

\section{Interleaving visual audio input}
\label{app:dual_context}
In the preceding experiments, we showed that omni-language models exhibit varying performance in pairwise cross-modal reasoning, particularly between vision–text and audio–text tasks. Yet, real-world multimodal scenarios are more complex: information from multiple modalities often arrives simultaneously and must be processed in an integrated manner. To address this challenge, we extend all tasks in XModBench to an audio–visual context configuration, where the question stem provides both audio and visual cues, while the candidate space remains identical to the original text-based setting.

We evaluate this dual-context setup using the Gemini series of models, which represent some of the most advanced omni-language systems available. The results, presented in Tab.~\ref{tab:dual_context}, enable a direct comparison with the pairwise baseline and reveal how models leverage—or fail to leverage—simultaneous multimodal evidence.

\begin{table}[!h]
\centering
\small
\caption{Overall performance of Gemini models under the dual-context setting 
(audio+visual context $\mapsto$ text). We compare with pairwise baselines (A $\mapsto$ T and V $\mapsto$ T), 
and report the stronger unimodal baseline $\max(\text{A $\mapsto$ T}, \text{V $\mapsto$ T})$.}
\small
\label{tab:dual_context}
\begin{tabular}{l|c|c|c}
\toprule
\textbf{Setting} & \textbf{Gemini 1.5 Pro} & \textbf{Gemini 2.0 Flash} & \textbf{Gemini 2.5 Pro}  \\
\midrule
~~~~~A $\mapsto$ T   & 52.76 & 63.71 & 70.99 \\
~~~~~V $\mapsto$ T   & 79.92 & 85.20 & 88.60 \\
A+V $\mapsto$ T & 82.53 (\textbf{+2.61}) & 79.84  &  89.76 \\

\bottomrule
\end{tabular}
\end{table}

\section{Human Survey}
\label{app:human}
\begin{figure}[b]
  \centering
  \includegraphics[width=0.7\linewidth]{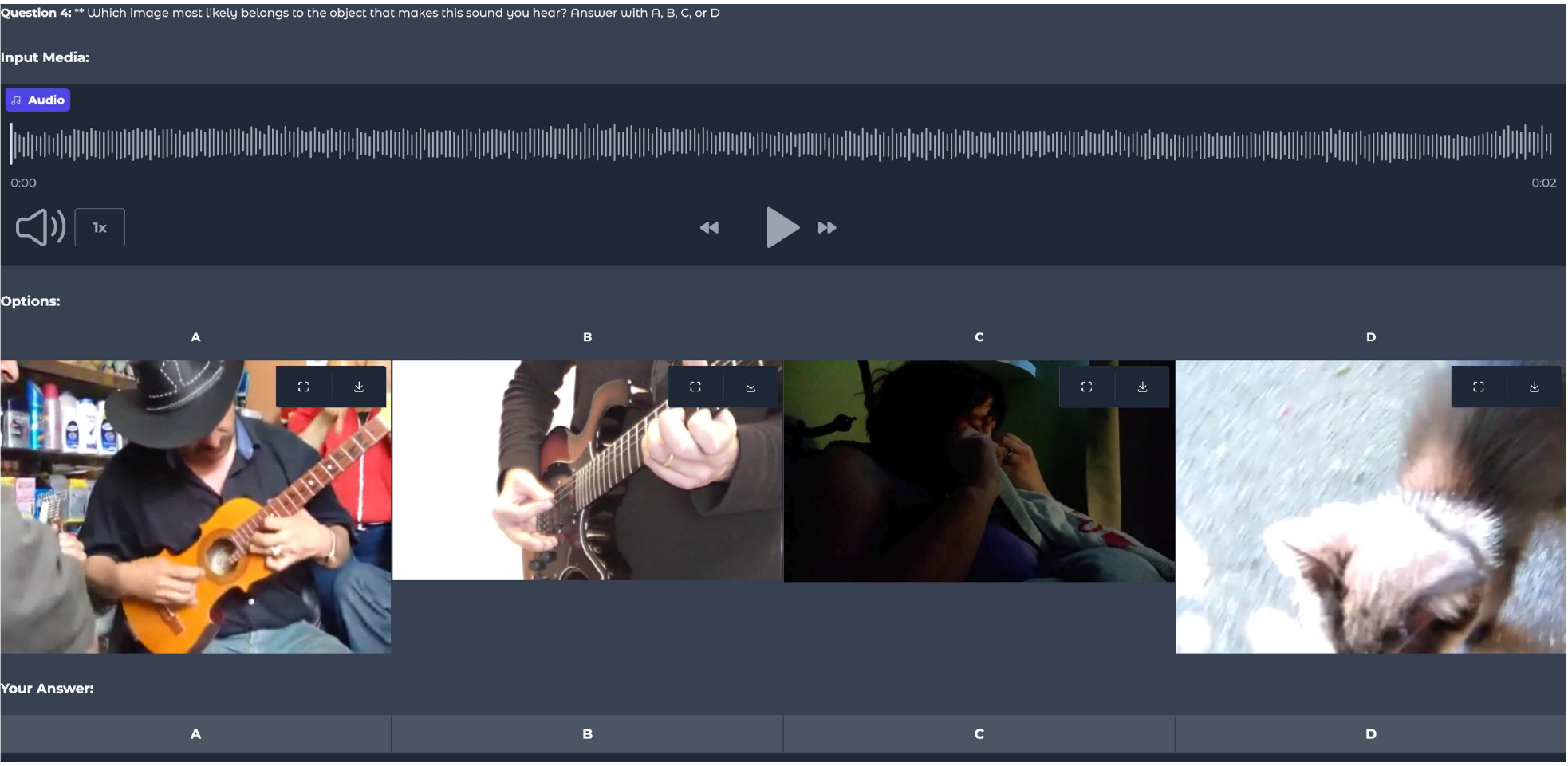}
  \caption{Sample question of human survey}
  \label{fig:human_survey}
\end{figure}

To evaluate human performance and establish reference baselines, we conducted a user study on a subset of \textbf{XModBench}. Participants answered multiple-choice questions under different modality configurations, with Figure~\ref{fig:human_survey} showing a screenshot of the interface and example questions. For each subtask, we collected responses from 10 valid participants per modality configuration.

\section{Techiniqal Details in 
\label{app:data}
Triplet Data collection and Processing Data  for Each Subtask}

In this section, we provide detailed descriptions of the data sources are collected, and how each data in each modality are processed for each subtask in XModBench. 

\subsection{Perceptual Recognition}\label{sec:data:perc}

\paragraph{General Categories.} 
We utilize the VGGSound Source (VGG-SS) dataset\citep{chen2021localizing,kim2024learning}, a large-scale video benchmark designed for sound source localization, which provides video-level annotations across diverse sound activities. The dataset covers 200 categories with approximately 5,000 video clips, where sound sources are annotated with bounding boxes to ensure clear visibility in each clip. For our benchmark, we extract a 2-second segment corresponding to the loudest audio channel as the audio input, and randomly sample a single frame from the same clip as the visual input. The activity class name serves as the textual description. To construct multiple-choice questions, four additional activity labels are randomly sampled as distractors, resulting in four candidate answers per instance. We then use Gemini 2.5-flash lite to\citep{comanici2025gemini} filter if each instance if the audio and video frame is clear to be hear and the image frame and audio are all match the category name.

\paragraph{Fine-grained Categories.} 
This subtask uses the same pool of video clips as the General Categories setting. The difference lies in reorganizing the activity classes into eight fine-grained clusters: \emph{Animal sounds}, \emph{Musical instruments}, \emph{Human activities}, \emph{Transportation}, \emph{Tools and utilities}, \emph{Urban sounds}, \emph{Human speech}, and \emph{Natural sounds}. For each instance, we select the target activity along with four distractor activities sampled from the same fine-grained cluster. This ensures that all answer choices belong to the same semantic domain, making the recognition task more challenging and diagnostic within a coherent category group.

\paragraph{Natural Environment.} 
We draw data from the Landscapes dataset\citep{lee2022sound}, which consists of ambient audio–video clips capturing natural outdoor scenes. Following the same selection protocol as in the General Categories task, we extract a 2-second segment from the dominant audio channel as the audio input, and randomly sample one frame from the corresponding video as the visual input. The dataset’s categorical labels are used as the textual descriptions.  

\paragraph{Instruments.} 
Instrument data is collected from the Solos dataset\citep{Montesinos2020SolosAD}, which contains recordings of 13 distinct instruments: violin, viola, cello, double bass, flute, oboe, clarinet, bassoon, saxophone, trumpet, horn, trombone, and tuba. We use the video frames as the visual modality, the isolated performance recordings as the audio modality, and the instrument names as textual labels.  

\paragraph{Instrument Composition.} 
We employ the URMP dataset\citep{li2018creating}, a multimodal corpus designed for music performance analysis, which provides video and audio recordings of ensemble performances. For this subtask, we leverage clips containing multiple instruments playing together, using the mixture audio as input, sampled video frames as the visual modality, and instrument combination labels as text.

\subsection{Spatial Reasoning}\label{sec:data:spat}

\paragraph{2D Horizontal Arrangement.} 
This subtask is derived from the URMP dataset\citep{li2018creating}, which contains multi-instrument ensemble recordings with annotated left-to-right spatial positions of each performer and independent audio channels per instrument. We construct multiple-choice questions by generating three distractor options through random shuffling of instrument order along the horizontal axis. For the visual modality, cropped player images are concatenated into a composite frame that preserves their spatial arrangement. For the audio modality, stereo spatialization is synthesized by assigning distinct azimuth values to each shuffled configuration and adjusting the relative channel balance using a panning algorithm (e.g., vector-base amplitude panning\citep{pulkki1997virtual}). This design ensures that listeners can clearly perceive the relative horizontal positions of the instruments.

\paragraph{3D Localization.}
This subtask builds on the STARSS23 dataset\citep{shimada2023starss23}, which provides panoramic video with time-stamped annotations of sound source depth, azimuth, and activity. For the visual modality, we annotate sound sources with bounding boxes and generate alternative views by rotating the camera perspective to $+90^\circ$, $180^\circ$, and $-90^\circ$ (positive defined as left). The corresponding videos are created through spatial cropping of frames. For the audio modality, we utilize the four-channel microphone array (MIC) recordings and simulate azimuthal rotation by first encoding the array signals into first-order Ambisonics (FOA), applying a 2D rotation matrix to the X–Y components, and decoding back into microphone signals with loudness normalization. To further enhance perceptual realism, each spatial microphone signal is additionally processed with head-related transfer functions (HRTFs) in the SOFA format\citep{majdak2013sofa, algazi2001cipic}.

\paragraph{3D Movements.}
This subtask is based on the Urbansas dataset\citep{fuentes2022urban}, which provides street-view traffic videos with detailed audio annotations indicating vehicle types and the presence of off-screen sounds. Each clip includes labels specifying the vehicle category, whether the sound source is visible in the video, and its temporal activity. We curate video segments from this dataset and highlight the target vehicle by overlaying a red bounding box to establish clear audio–visual correspondence.

\subsection{Temporal Reasoning}\label{sec:data:temp}

\paragraph{Event Order.}
This subtask is derived from 2-second video clips in the VGGSound Source (VGG-SS) dataset\citep{chen2021localizing}, originally used in the Perceptual Recognition task where each clip is annotated with an activity class label. For temporal ordering, we randomly sample 3–5 clips from different classes and generate four candidate event sequences by shuffling their order. Each sequence is represented across three modalities: (i) a text description (e.g., ``Event A $\rightarrow$ Event B $\rightarrow$ Event C''), (ii) a concatenated video sequence, and (iii) a concatenated audio sequence. Multiple-choice questions are formed by selecting one sequence as the correct answer and presenting the stem in one modality, while the four candidate sequences are given in another modality.

\paragraph{Repetition Count.}
Following the setup in\citep{zhang2021repetitive}, this subtask focuses on counting repeated events. Visual data is generated from synthetic renderings of repeated object actions, while audio data consists of temporal patterns with clear repetitions (e.g., sequences of knocks or claps). Text prompts explicitly query the number of repetitions in either modality.

\paragraph{Repetition Calculation.}
Also inspired by\citep{zhang2021repetitive}, this subtask extends beyond direct counting by requiring simple arithmetic over observed repetitions. Both audio and video are rendered with variable frequencies of repeated events, while the text prompts encode arithmetic formulations that ask models to compute totals (e.g., ``three knocks plus two knocks'').

\subsection{Linguistic Understanding}\label{sec:data:ling}

\paragraph{Linguistic Recognition.}
This subtask targets recognition of textual content across modalities. Images are collected from OCR-rendered text data\citep{RenderedText}, each paired with its ground-truth transcript. Audio is generated from these transcripts using a TTS system\citep{guo2025fireredtts}, allowing for cross-modal recognition between text, vision, and speech.

\paragraph{Translation.}
This subtask examines cross-lingual translation. Input sequences consist of English text with multiple-choice options in Chinese. Text data is derived from OCR-rendered images\citep{RenderedText}, while translations are generated using Gemini\citep{team2024gemini}. Visual inputs are rendered using the OCR dataset rendering toolkit\citep{gbothq_ocr_dataset_rendering}, and audio is synthesized from both languages with a TTS system\citep{guo2025fireredtts}.

\paragraph{Dialogue Emotion.}
This subtask focuses on multimodal emotion recognition in conversational settings. Visual data consists of face videos displaying emotional expressions extracted from multi-party dialogue clips\citep{chen2018emotionlines, poria2018meld}. Each dialogue is paired with transcripts and annotated with categorical emotions (anger, disgust, fear, sadness, surprise, and joy). We filter clips to lengths between 5–30 seconds. The video data is stripped of original audio but accompanied by transcripts to enable inference of emotion from dialogue and facial expression. Audio inputs consist of the original speech tracks, and text inputs are provided as the emotion category names.

\subsection{External Knowledge}\label{sec:data:knwl}
\paragraph{Music Genre Classification.}
This subtask evaluates music genre recognition. We collect audio samples from the GTZAN dataset\citep{olteanu_gtzan_kaggle}, covering multiple musical styles. To complement the audio, we also collect representative album cover images for each genre category.

\paragraph{Movie Matching.}
This subtask requires linking multimodal cues to movie identities. We collect a set of recent films from IMDb. For the visual modality, we use official posters. To prevent trivial text matching between posters and movie titles, we use written plot summaries from IMDb as the text modality. Audio is sampled as 30-second clips from publicly available trailers on YouTube.

\paragraph{Singer Identification.}
This subtask targets cross-modal recognition of popular singers. Images of singers are collected from the web, while audio consists of short clips (3–5 songs each) sampled from their publicly available music videos on YouTube. Text inputs include singer names and associated biographical metadata. We select a diverse set of internationally recognized artists, including American singers Ariana Grande, Bad Bunny, Billie Eilish, Bruno Mars, Chappell Roan, Harry Styles, and Chinese singers David Tao, Eason Chan, Faye Wong, G.E.M., and Jay Chou.

% \section{LLM Usage}
% In this work, large language models were employed for grammar refinement and language polishing. In addition, we utilized LLM to filter and refine the benchmark data. All substantive contributions—including the design of the conceptual framework, experimental studies, and the writing of technical content—are entirely original and carried out by the authors.

\section{LLM Usage}\label{sec:llm}
We used large language models (LLMs) to assist in the preparation of this paper. Their role was limited to language editing such as proofreading and rephrasing. All ideas, experiments, and analyses were conceived and conducted by the authors.

\end{document}